%% file: main.tex
\documentclass[10pt,twocolumn,letterpaper]{article}

\usepackage{cvpr}              % To produce the CAMERA-READY version

\definecolor{cvprblue}{rgb}{0.21,0.49,0.74}

\usepackage{natbib}
\setcitestyle{numbers,square}
\usepackage[utf8]{inputenc} 
\usepackage[T1]{fontenc}     
\usepackage{url}

\usepackage[toc,page,header]{appendix}
\usepackage{minitoc}

\usepackage{booktabs}       
\usepackage{amsfonts}       
\usepackage{amsmath}
\usepackage{amssymb}
\usepackage{mathtools}
\usepackage{amsthm}
\usepackage{nicefrac}       
\usepackage{microtype}      
\usepackage{xcolor}         
\usepackage{graphicx}
\usepackage{subcaption}
\usepackage{multirow}
\usepackage{makecell}
\usepackage{algorithm}
\usepackage{algorithmic}
\usepackage{threeparttable}
\theoremstyle{plain}
\newtheorem{theorem}{Theorem}[section]

\theoremstyle{definition}
\newtheorem{definition}[theorem]{Definition}

\theoremstyle{remark}

\theoremstyle{plain}

\usepackage[pagebackref,breaklinks,colorlinks,allcolors=cvprblue]{hyperref}
\usepackage[accsupp]{axessibility}

\def\paperID{14084} % *** Enter the Paper ID here
\def\confName{CVPR}
\def\confYear{2025}

\title{UniAP: Unifying Inter- and Intra-Layer Automatic Parallelism by Mixed Integer Quadratic Programming}

\author{  Hao Lin\footnotemark[1],~~Ke Wu\footnotemark[1],~~Jie Li\footnotemark[1],~~Jun Li,~~Wu-Jun Li\footnotemark[2]\\
  National Key Laboratory for Novel Software Technology \\
  School of Computer Science \\
  Nanjing University, Nanjing 210023, China \\
  \texttt{\{}\href{mailto:hao.lin@smail.nju.edu.cn}{\texttt{hao.lin}}\texttt{,} \href{mailto:ke.wu@smail.nju.edu.cn}{\texttt{ke.wu}}\texttt{,} \href{mailto:jie-li@smail.nju.edu.cn}{\texttt{jie-li}}\texttt{,} \href{mailto:lijun@smail.nju.edu.cn}{\texttt{lijun}}\texttt{\}@smail.nju.edu.cn}\texttt{,} \href{mailto:liwujun@nju.edu.cn}{\texttt{liwujun@nju.edu.cn}}
}

\begin{document}
\include{symbol_def}
\maketitle
\renewcommand{\thefootnote}{\fnsymbol{footnote}}
\footnotetext[1]{Equal contribution.}
\footnotetext[2]{Corresponding author.\par
}

\renewcommand{\thefootnote}{\arabic{footnote}}
\input{sec/0_abstract}    
\input{sec/1_intro}
\input{sec/2_background}

\input{sec/3_method}

\input{sec/4_experiments}
\input{sec/5_conclusion}
\input{sec/7_acknowledgement}
{
    \small
    \bibliographystyle{ieeenat_fullname}
    \bibliography{main}
}

\input{sec/X_suppl}

\end{document}

%% file: sec/0_abstract.tex
\begin{abstract}
 Distributed learning is commonly used for training deep learning models, especially large models. In distributed learning, manual parallelism~(MP) methods demand considerable human effort and have limited flexibility. Hence, automatic parallelism~(AP) methods have recently been proposed for automating the parallel strategy optimization process. Existing AP methods suffer from sub-optimal solutions because they do not jointly optimize the two categories of parallel strategies~(i.e., inter-layer parallelism and intra-layer parallelism). In this paper, we propose a novel AP method called UniAP, which unifies inter- and intra-layer automatic parallelism by mixed integer quadratic programming. To the best of our knowledge, UniAP is the first parallel method that can jointly optimize the two categories of parallel strategies to find an optimal solution. Experimental results show that UniAP outperforms state-of-the-art methods by up to 3.80$\times$ in throughput and reduces strategy optimization time by up to 107$\times$ across five Transformer-based models.
\end{abstract}

%% file: sec/1_intro.tex
\section{Introduction}
\label{sec:introduction}
% Deep learning models have demonstrated promising performance across many domains. 
% For example, deep learning models such as BERT~\citep{devlin_bert_2019}, GPT-3~\citep{brown_language_2020} and T5~\citep{raffel_exploring_2020} achieve state-of-the-art~(SOTA) performance on many natural language processing~(NLP) tasks. 
% For computer vision~(CV), deep learning models such as ViT~\citep{dosovitskiy_image_2021} and Swin Transformer~\citep{liu_swin_2021} achieve good accuracy on multiple tasks. 

Distributed learning~(also called parallel learning) on clusters with several machines or GPUs is commonly used for training deep learning models, especially for some large models with billions of parameters~\citep{brown_language_2020,touvron_llama_2023,touvron_llama_2023-1}. Several parallel strategies, including pipeline parallelism~(PP), data parallelism~(DP), tensor parallelism~(TP), and fully sharded data parallelism~(FSDP), have been proposed for distributed learning. These parallel strategies can be divided into two main categories: inter-layer parallelism and intra-layer parallelism. Inter-layer parallelism~\citep{huang_gpipe_2019,narayanan_pipedream_2019,narayanan_memory-efficient_2021,fan_dapple_2021,li_chimera_2021,lepikhin_gshard_2021,du_glam_2022,fedus_switch_2022}, which includes PP, partitions the model into disjoint sets without partitioning tensors in each layer. 
Intra-layer parallelism~\citep{li_pytorch_2020,rasley_deepspeed_2020,narayanan_efficient_2021,fairscale_authors_fairscale_2021}, which includes DP, TP, and FSDP, partitions tensors in a layer along one or more axes.

The parallel method\footnote{To avoid confusion, we treat `parallel method' and `parallel strategy' as two different terminologies in this paper.} in one specific distributed learning method or system typically adopts one parallel strategy or a combination of several parallel strategies. 
Existing parallel methods can be divided into two categories: manual parallelism~(MP) methods and automatic parallelism~(AP) methods. %Different combinations of parallel strategies will significantly affect the training efficiency of the    distributed learning process. To enhance the training efficiency, researchers have proposed some manually designed parallel training strategies~\citep{narayanan_efficient_2021,shazeer_mesh-tensorflow_2018,xu_gspmd_2021}. 
In MP methods ~\citep{shazeer_mesh-tensorflow_2018,narayanan_efficient_2021,xu_gspmd_2021}, one or several parallel strategies are manually optimized by researchers or developers. MP methods require extensive domain knowledge in deep learning models and hardware architectures. With the rapid development of deep learning models and the increasing diversity of modern hardware architectures~\citep{flynn_very_1966,flynn_computer_1972}, MP methods demand considerable human effort and have limited flexibility.

To address the two limitations of MP methods, AP methods~\citep{narayanan_pipedream_2019,he_pipetransformer_2021,zheng_alpa_2022} have recently been
proposed for automating the parallel strategy optimization process. Although existing AP methods have achieved promising progress, they optimize the two categories of parallel strategies separately rather than jointly. 
More specifically, some methods optimize only one category of parallel strategies~\citep{jia_exploring_2018,wang_supporting_2019,jia_beyond_2019,schaarschmidt_automap_2021,zhao_vpipe_2022,cai_tensoropt_2022,liu_colossal-auto_2023}, and the others optimize inter- and intra-layer parallelism hierarchically~\citep{narayanan_pipedream_2019,tarnawski_efficient_2020,tarnawski_piper_2021,fan_dapple_2021,he_pipetransformer_2021,zheng_alpa_2022}. Hence, existing AP methods suffer from sub-optimal solutions.

%We hereby summarize our contributions as follows:
In this paper, we propose a novel AP method called UniAP for distributed learning. The contributions of UniAP are outlined as follows: 
\begin{itemize}
\item 
 UniAP unifies inter- and intra-layer automatic parallelism by mixed integer quadratic programming~(MIQP)~\citep{lazimy_mixed_1982}. 
 \item To the best of our knowledge, UniAP is the first parallel method that can optimize the two categories of parallel strategies jointly rather than separately to find an optimal solution.
 \item Experimental results show that UniAP outperforms state-of-the-art methods by up to 3.80$\times$ in throughput and reduces strategy optimization time by up to 107$\times$ across five Transformer-based models.
\end{itemize}

%% file: sec/2_background.tex
\section{Background}
% \subsection{Inter- and Intra-Layer Parallelism}
% In general, there exist two main categories of parallel strategies for deep learning models: inter- and intra-layer parallelism. If we want to divide them further, inter-layer parallelism mainly includes pipeline parallelism~(PP) in our context. Meanwhile, intra-layer parallelism mainly includes data parallelism~(DP), tensor parallelism~(TP), and fully sharded data parallelism~(FSDP). Most MP and AP methods optimize for the optimal parallel strategy within these dimensions.
\subsection{Parallel strategy}
\label{subsec:background:parallel-strtegy}
\textbf{Pipeline parallelism~(PP)}\quad In PP~\citep{huang_gpipe_2019}, each worker~(machine or GPU) holds a subset of model layers. Adjacent layers on different workers need to transfer activations in the forward propagation~(FP) step and gradients in the backward propagation~(BP) step. 
%UniAP focuses on synchronous PP, which performs weight updating in each stage at the end of each iteration.

\noindent\textbf{Data parallelism~(DP)}\quad In DP~\citep{li_pytorch_2020}, each worker holds a replica of the whole model and partitions training samples. In each iteration, each worker computes gradients and synchronizes them with the other workers using all-reduce collective communication~(CC). All workers will have the same model parameters after the synchronization step.

\noindent\textbf{Tensor parallelism~(TP)}\quad In TP~\citep{narayanan_efficient_2021}, each worker holds a replica of training samples and partitions within model layers. In each iteration, each worker computes its local outputs in FP and its local gradients in BP. To synchronize outputs and gradients, all workers will perform all-reduce CC in FP and BP steps according to the partition scheme.

\noindent\textbf{Fully sharded data parallelism~(FSDP)}\quad FSDP~\citep{rajbhandari_zero_2020,fairscale_authors_fairscale_2021} partitions optimizer states, parameters and gradients of the model into separate workers. During the FP and BP step of each iteration, FSDP performs an all-gather CC to obtain the complete parameters for the relevant layer, respectively. After computing gradients, FSDP conducts a reduce-scatter CC to distribute the global gradients among the workers.

\subsection{Parallel method}
\textbf{Manual parallelism~(MP)}\quad
MP refers to the parallel methods in which human experts design and optimize the parallel strategies. Representative MP methods include Megatron-LM~\citep{narayanan_efficient_2021}, Mesh-TensorFlow~\citep{shazeer_mesh-tensorflow_2018}, and GSPMD~\citep{xu_gspmd_2021}. Megatron-LM manually designs TP and PP strategies for training Transformer-based models and exhibits superior efficiency. Mesh-TensorFlow and GSPMD require human effort to designate and tune the intra-layer parallel strategy. These methods rely on expert design and have little flexibility, challenging their automatic application to other models.

\noindent\textbf{Inter-layer-only AP or intra-layer-only AP}\quad For inter-layer-only AP, GPipe~\citep{huang_gpipe_2019} and vPipe~\citep{zhao_vpipe_2022} employ a balanced partition algorithm and a dynamic layer partitioning middleware to partition pipelines, respectively. For intra-layer-only AP, OptCNN~\citep{jia_exploring_2018}, TensorOpt~\citep{cai_tensoropt_2022}, and Tofu~\citep{wang_supporting_2019} employ dynamic programming methods to optimize DP and TP strategies together. FlexFlow~\citep{jia_beyond_2019} and Automap~\citep{schaarschmidt_automap_2021} use the Monte Carlo method to find the optimal DP and TP strategy. Colossal-Auto~\citep{liu_colossal-auto_2023} utilizes integer programming~(IP) techniques to generate intra-layer parallelism and activation checkpointing strategies. All these methods optimize only one category of parallel strategies.

\noindent\textbf{Inter- and intra-layer AP}\quad PipeDream~\citep{narayanan_pipedream_2019}, DAPPLE~\citep{fan_dapple_2021}, and PipeTransformer~\citep{he_pipetransformer_2021} use dynamic programming to determine optimal strategies for both DP and PP. DNN-partitioning~\citep{tarnawski_efficient_2020} adopts IP and dynamic programming to explore DP and PP strategies. DT-FM~\citep{yuan_decentralized_2022} combines genetic algorithm and the local search strategy to explore DP and PP strategies in a layer-wise manner. Piper~\citep{tarnawski_piper_2021} and Alpa~\citep{zheng_alpa_2022} adopt a parallel method considering DP, TP, and PP.
Galvatron~\citep{miao_galvatron_2022} uses dynamic programming to determine DP, TP, and FSDP strategies in a single pipeline stage. As for PP, it partitions stages and determines micro-batch size using naive greedy algorithms. All these methods are hierarchical, which will result in sub-optimal solutions.

%% file: sec/3_method.tex
\begin{figure}[t]
\begin{center}
\centerline{\includegraphics[width=\linewidth]{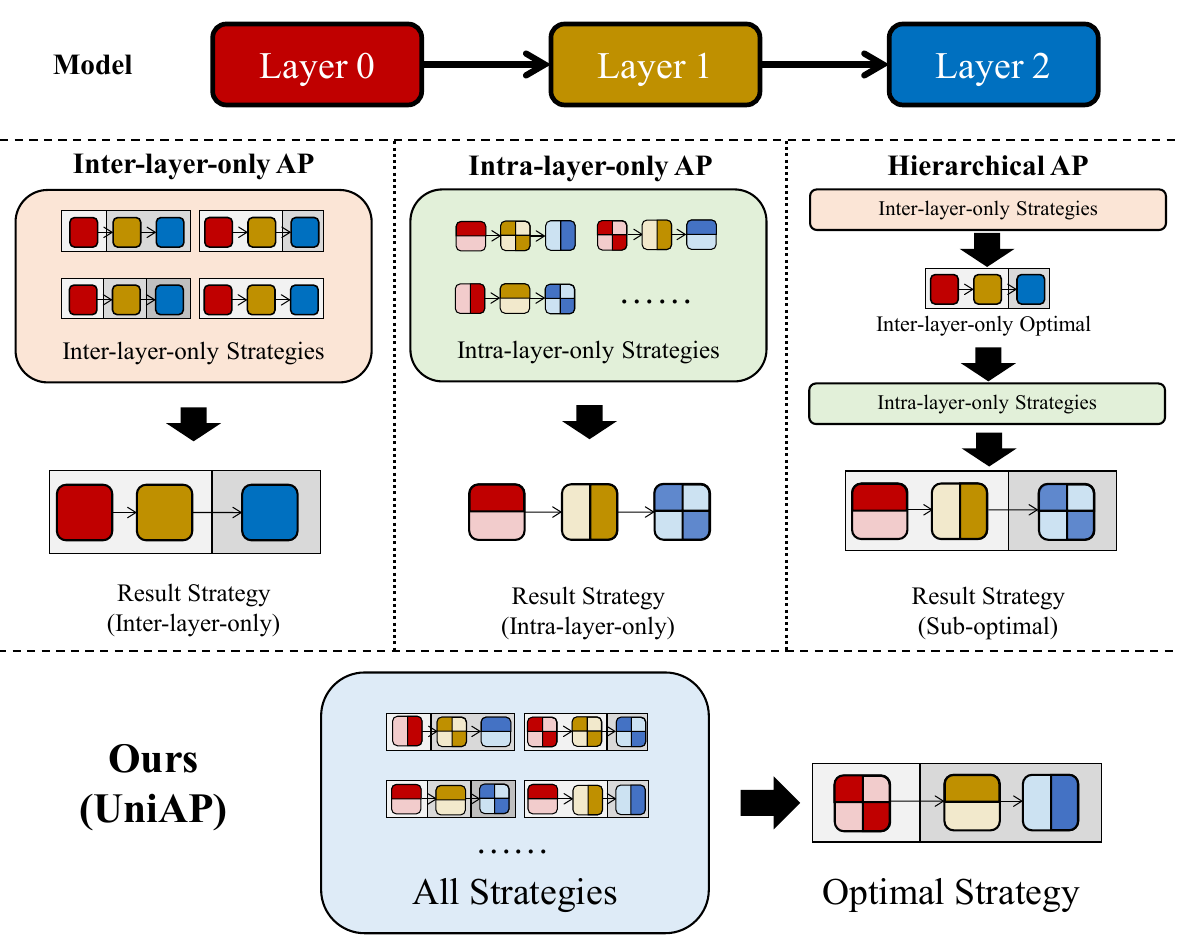}}
\caption{Parallel methods for optimizing parallel strategies for a three-layer model. The 
different arrangements of slices with varying transparency within the same layer block indicate different intra-layer parallelism strategies adopted by layers. The different arrangements of gray blocks which wrap the layer blocks indicate different inter-layer parallelism strategies.}
\label{fig:inter_intra_AP}
\end{center}
\vskip -0.4in
%\vskip -0.2in
\end{figure}

\section{Method}
\label{sec:method}
In this section, we introduce our proposed method called UniAP, which jointly optimizes the two categories of
parallel strategies, including PP, DP, TP, and FSDP, to find an optimal solution. Figure~\ref{fig:inter_intra_AP} illustrates the difference between UniAP and other automatic parallelism methods. Inter-layer-only and intra-layer-only AP methods optimize~(search) from a set of candidate inter-layer-only and intra-layer-only parallel strategies, respectively. 
Hierarchical AP methods first adopt greedy or dynamic programming to propose candidate inter-layer parallel strategies. Then, they optimize the intra-layer parallel strategy for every fixed inter-layer parallel strategy. 
UniAP has the largest strategy space for exploration~(joint optimization).
% Figure~\ref{fig:parallel_methods} illustrates the difference between UniAP and other AP methods. Inter-layer-only and intra-layer-only AP methods optimize~(search) from a set of candidate inter-layer-only and intra-layer-only parallel strategies, respectively. 
% Hierarchical AP methods first adopt greedy or dynamic programming to propose candidate inter-layer parallel strategies. Then, they optimize the intra-layer parallel strategy for every fixed inter-layer parallel strategy. 
% Apart form them, UniAP has the largest strategy space for exploration~(joint optimization).
%Hierarchical AP methods first optimize the inter-layer parallel strategy, and then optimize the intra-layer parallel strategy after the inter-layer parallel strategy has been optimized and fixed.

\begin{figure}[t]
\begin{center}
\centerline{\includegraphics[width=\linewidth]{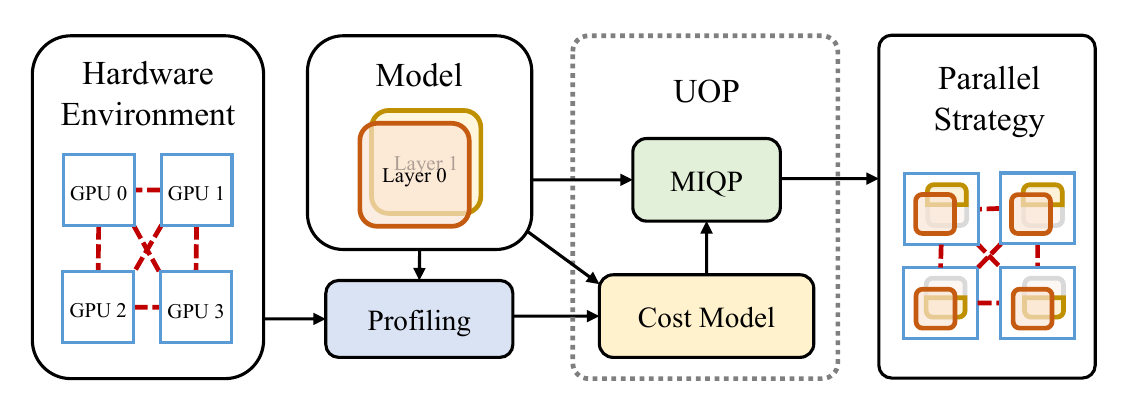}}
\caption{Flowchart of UniAP.}
\label{fig:overview}
\end{center}
\vskip -0.4in
\end{figure}

Figure~\ref{fig:overview} illustrates the flowchart of UniAP. UniAP first profiles the runtime information for the user's hardware environment and the deep learning model. After that, UniAP estimates inter- and intra-layer costs given the computation graph and profiling results with its cost models. The estimated costs and the computation graph are then transformed into an MIQP problem. The objective function of the MIQP is to maximize the training throughput, or in other words, to minimize the training time per iteration (TPI). By iteratively applying the cost model and MIQP with different parameters, UniAP determines the minimal TPI and its corresponding parallel strategies. We name this process the Unified Optimization Process~(UOP). Finally, UniAP interprets the parallel strategies into the execution plan for the designated model.

\subsection{Profiling}\label{subsec:method:profiling}
UniAP collects runtime information about the hardware environment and deep learning model during profiling. For the hardware environment, UniAP evaluates the efficiency of all-reduce and point-to-point (P2P) communication for different device subsets. For example, when profiling a node with 4 GPUs, UniAP measures the all-reduce efficiency for various DP, TP, and FSDP combinations across these GPUs. Additionally, UniAP ranks these GPUs from 0 to 3 and evaluates the speed of P2P for two pipeline options: ($0\rightarrow 2$ and $1\rightarrow 3$) and ($0\rightarrow 1$, $1\rightarrow 2$ and $2\rightarrow 3$). Furthermore, UniAP estimates the computation-communication overlap coefficient~(CCOC)~\citep{miao_galvatron_2022,rashidi_enabling_2021}.

UniAP acquires two types of information for the deep learning model: computation time and memory usage. On one hand, UniAP distinguishes the forward computation time per sample for different types of hidden layers. On the other hand, UniAP collects memory usage information for each layer, including the memory occupied by parameters and the memory usage of activation per sample in different TP sizes. 

\subsection{Cost model}\label{subsec:method:cost-model}
UniAP employs two primary cost models, namely the time cost model and the memory cost model. 

\noindent\textbf{Time cost model}\quad To estimate computation time, UniAP first calculates the forward computation time by multiplying the batch size with the forward computation time per sample obtained from profiling. Users can obtain a more precise result by profiling the forward time on the specified batch size. For Transformer-based models that mainly consist of the MatMul operator, the computation time in the BP stages is roughly twice that of the FP stages~\citep{narayanan_efficient_2021,li_chimera_2021,miao_galvatron_2022}. Additionally, UniAP estimates the communication time by dividing the size of transmitting tensors by the profiled communication efficiency for different communication primitives. To accommodate overlapping, UniAP multiplies the profiled CCOC by the overlapping interval of computation and communication. To model the communication time between pipeline stages, UniAP calculates the cross-stage cost between consecutive stages by the summation of P2P costs.

\noindent\textbf{Memory cost model}\quad UniAP estimates memory consumption for each layer with its memory cost model. This estimation consists of three steps for a given layer. First, it computes the activation memory cost $m_a$ by multiplying the batch size and the profiled activation memory cost per sample of the TP size used by the strategy. Next, UniAP calculates the memory cost of model states $m_{s}$ for each layer based on their parameter size $ps$, TP size $ts$, FSDP size $fs$, and a constant $c_{dtype}$ dependent on the data type. Formally, we have 
\begin{equation}
m_s=\frac{c_{dtype}\times ps}{ts\times fs}.
\end{equation}
For example, if we choose precision as FP32, then $c_{dtype}=(4+4+4+4)/4=4$ since the learnable parameters, gradients, momentum and variance consume equal size of memory. If we opt for mixed precision with FP16 activated, then $c_{dtype}=(4+4+4+2+2)/2=8$. Finally, UniAP aggregates the activation memory cost $m_a$, memory cost of model states $m_s$, and context memory cost $m_c$ to a constant matrix $\textbf{\textit{M}}$, where $\textbf{\textit{M}}_{uk}$ denotes the memory cost for the $k$-th intra-layer strategy of layer $u$ on a single device.
% Overall, these two cost models employed by UniAP strike a balance between complexity and accuracy.

\subsection{Mixed integer quadratic programming}\label{subsec:method:miqp}
%This section describes our MIQP expression in terms of a formulation-oriented approach. We will firstly elaborate the objective function, and then move on to the constraints.
% The estimated costs, along with the computation graph, are then transformed into an MIQP problem, the formulation of which  includes an objective function and several constraints.
The estimated costs and the computation graph are then transformed into an MIQP problem. Its formulation includes an objective function and several constraints.

\subsubsection{Objective function}\label{subsubsec:method:objective-function}
\begin{figure*}[t]
\begin{minipage}[t]{0.6\linewidth}
    \centering
    \includegraphics[width=\linewidth]{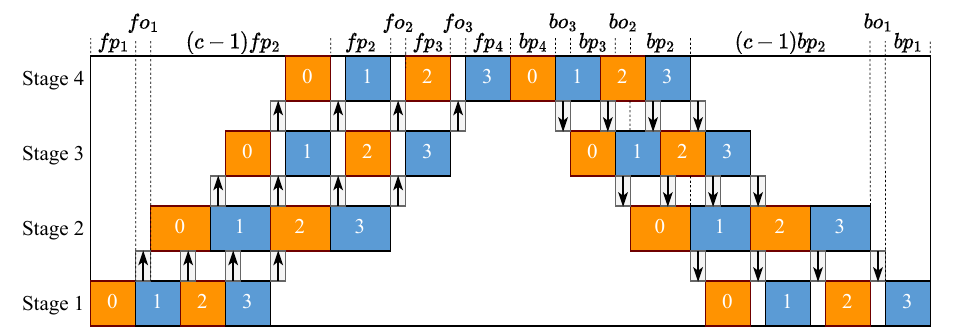}
    \caption{Time cost decomposition of a GPipe-style PP.}
\label{fig:gpipe}
\end{minipage}
\hskip -0.9in
\begin{minipage}[t]{0.66\linewidth}
    \centerline{\includegraphics[width=0.58\linewidth]{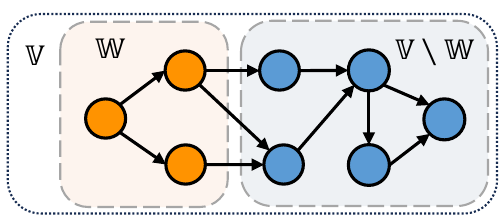}}
    \caption{A contiguous set.}
    \label{fig:contiguous}
\end{minipage}
\vskip -0.1in
\end{figure*}
The objective function tries to minimize TPI. In this paper, we have chosen GPipe as our PP strategy for illustration.\footnote{UniAP is also compatible with other PP strategies. For example, users need to modify only the memory constraint in Section \ref{subsubsec:method:constraints} to adapt to synchronous 1F1B pipeline~\citep{narayanan_memory-efficient_2021,fan_dapple_2021}.} Figure~\ref{fig:gpipe} depicts the time cost decomposition of a GPipe-style PP with non-negligible communication costs. The time needed to apply gradients at the end of each iteration is not included, as it depends on the optimizer and is insignificant compared to the total time spent on FP and BP.

We denote the cost for computation stages as $\mathbb{P}=\{p_1, p_2, \dots, p_{deg}\}$ and the cost for communication stages as $\mathbb{O}=\{o_1, o_2, \dots, o_{deg-1}\}$. Here, $deg$ represents the number of computation stages, which corresponds to the pipeline parallel size. In Figure~\ref{fig:gpipe}, $fp_i$ and $bp_i$ denote forward and backward computation time for computation stage $i$, respectively. $fo_j$ and $bo_j$ denote forward and backward communication time for communication stage $j$, respectively. Hence, we have $p_i=fp_i+bp_i$ and $o_j=fo_j+bo_j$.

In a GPipe-style pipeline, we use $c$ to denote the number of micro-batches. As illustrated in Figure \ref{fig:gpipe}, a mini-batch is uniformly split into four micro-batches, and the total TPI is determined by the latency of all computation and communication stages and the latency of the slowest stage. We further denote TPI in GPipe as $tpi_{gpipe}$. Given that a stage with a higher FP computation cost leads to a higher BP computation cost with high probability, we can write the objective function of GPipe-style pipeline as follows:
\begin{equation}
\begin{aligned}
\min tpi_{gpipe}&=\sum_{i=1}^{deg} p_i + \sum_{j=1}^{deg-1} o_{j} \\ &+ (c - 1)\max\left(\mathbb{P}\cup \mathbb{O}\right). 
\end{aligned} 
\label{eqn:method:lp-contig}   
\end{equation}
\subsubsection{Constraint}\label{subsubsec:method:constraints}
We first introduce additional notations before presenting the constraints. For a given layer $u \in \mathbb{V}$, $\mathbb{S}_u$ represents its set of intra-layer parallel strategies, $\textbf{\textit{A}}_{uk}$ denotes the $k$-th intra-layer execution cost obtained from our time cost model. Additionally, we use $\textbf{\textit{S}}_{uk} \in \{0,1\}$ to indicate whether the $k$-th parallel strategy is selected for the layer $u$, and use $\textbf{\textit{P}}_{ui}\in \{0,1\}$ to indicate whether layer $u$ is placed on the $i$-th computation stage. Each edge $\langle u,v\rangle\in \mathbb{E}$ is assigned a resharding cost denoted by $\textbf{\textit{R}}_{uv}$ if the vertices are located within the same pipeline stage. Alternatively, if the vertices are located across consecutive stages, the resharding cost between them is denoted by $\textbf{\textit{R}}'_{uv}$. These two resharding costs are constant matrices derived from our time cost model.

%Subsequently, we proceed to the constraints of the MIQP. 

\noindent\textbf{Computation-stage constraint}\quad To compute the total cost for a single computation stage $i$, all computation and communication costs associated with that stage must be aggregated and assigned to $p_i$. This constraint can be formulated as follows:
\begin{equation}
\begin{aligned}
    \sum_{u\in \mathbb{V}}\textbf{\textit{P}}_{ui}\textbf{\textit{S}}_{u}^\mathsf{T}\textbf{\textit{A}}_{u}+\sum_{\langle u,v\rangle\in \mathbb{E}}\textbf{\textit{P}}_{ui}\textbf{\textit{P}}_{vi}(\textbf{\textit{S}}_{u}^\mathsf{T}\textbf{\textit{R}}_{uv}\textbf{\textit{S}}_{v})=p_i, \\
    ~\forall i\in\{1,\dots, deg\}.
\end{aligned}
\label{eqn:method:intra-stage}
\end{equation}
On the left side of Equation~\eqref{eqn:method:intra-stage}, the first polynomial term represents the cost of choosing specific intra-layer strategies for layers placed in stage $i$. The second term represents total resharding costs within stage $i$.

\noindent\textbf{Communication-stage constraint}\quad To calculate the total cost for a single communication stage $j$, we should aggregate the P2P costs incurred between consecutive stages and assign them to $o_j$. This constraint can be formulated as follows:
\begin{equation}
\begin{aligned}
    &~&\sum_{\langle u,v\rangle\in \mathbb{E}}\textbf{\textit{P}}_{uj}\textbf{\textit{P}}_{v(j+1)}(\textbf{\textit{S}}_{u}^\mathsf{T}\textbf{\textit{R}}'_{uv}\textbf{\textit{S}}_{v})=o_j,\\
    &~&~\forall j\in\{1,\dots,deg-1\}.
\end{aligned}
\label{eqn:method:inter-stage}
\end{equation}

\noindent\textbf{Memory constraint}\quad We need to guarantee that no devices~(GPUs) will encounter out-of-memory~(OOM) exceptions during training process. This constraint can be formulated as follows:
\begin{equation}
\sum_{u\in \mathbb{V}}\textbf{\textit{P}}_{ui} \textbf{\textit{S}}_u^\mathsf{T} \textbf{\textit{M}}_u\leqslant m,~\forall i\in\{1,\dots,deg\}.
\label{eqn:method:memory-constraint}
\end{equation}
Here, $m$ denotes the memory limit for each device. In the case of homogeneous computing devices, the value of $m$ remains constant throughout all stages. But the value of $m$ varies in the case of heterogeneous computing devices.

\noindent\textbf{Order-preserving constraint}\quad PP is not a single-program multiple-data~(SPMD) parallel strategy~\citep{huang_gpipe_2019}. Hence, we need an order-preserving constraint to ensure that the subgraphs of $\mathcal{G}$ are contiguous. We adopt the definition of \emph{contiguous} from previous work~\citep{tarnawski_efficient_2020,tarnawski_piper_2021}. 
\begin{definition}\label{def:contiguous}
A set $\mathbb{W} \subseteq \mathbb{V}$ is contiguous if there do not exist nodes $u \in \mathbb{W}$, $v \in \mathbb{V} \setminus \mathbb{W}$, and $w \in \mathbb{W}$ such that $v$ is reachable from $u$ and $w$ is reachable from $v$.
\end{definition}
% \begin{figure}[t]
% \vskip 0.2in
% \begin{center}
% \centerline{\includegraphics[width=0.5\linewidth]{contiguous-set.pdf}}
% \caption{A contiguous set $\mathbb{W}$.}
% \label{fig:contiguous}
% \end{center}
% \vskip -0.2in
% \end{figure}

Figure~\ref{fig:contiguous} illustrates an example of a contiguous set $\mathbb{W}$, in which we cannot find any reachable node pairs $\langle u,v\rangle$ and $\langle v,w\rangle$ where $u,w\in \mathbb{W}$ and $v \in \mathbb{V} \setminus \mathbb{W}$. 

In our case, our model will not be assigned to different pipeline stages in a disordered manner if we ensure that all subgraphs on each computation stage are contiguous. After reformulating this constraint in linear form, we have
\begin{subequations}
\label{eqn:method:order-preserving}
\begin{align}
    &\textbf{\textit{Z}}_{vi}\geqslant \textbf{\textit{P}}_{vi},&\notag\\&~\forall v\in \mathbb{V},~\forall i\in\{1,2,\dots,deg\},\label{eqn:method:order-preserving:1}\\
    &\textbf{\textit{Z}}_{vi}\leqslant \textbf{\textit{Z}}_{ui},&\notag\\&~\forall u,v\in \mathbb{V},~\forall \langle u,v\rangle\in \mathbb{E},~\forall i\in\{1,2,\dots,deg\},\label{eqn:method:order-preserving:2}\\
    &\textbf{\textit{Z}}_{vi}\leqslant \textbf{\textit{P}}_{vi}-\textbf{\textit{P}}_{ui} +1,&\notag\\&~\forall u,v\in \mathbb{V},~\forall \langle u,v\rangle\in \mathbb{E},~\forall i\in\{1,2,\dots,deg\}.\label{eqn:method:order-preserving:3}
\end{align}
\end{subequations}
Here, $\textbf{\textit{Z}}$ is an auxiliary variable shaped like $\textbf{\textit{P}}$. Given the node set $\mathbb{V}_i$ that contains nodes in stage $i$, we define $\textbf{\textit{Z}}_{vi}=1$ if a node $w\in \mathbb{V}_i$ is reachable from $v~(v\in\mathbb{V})$. Otherwise, $\textbf{\textit{Z}}_{vi}=0$.
Detailed proof can be found in Appendix~\ref{appendix:linear-form-of-contigous-constraint}.

\noindent\textbf{Layer-placement constraint}\quad All layers should be placed on exactly one pipeline stage and at least one layer should be placed on each pipeline stage. This constraint can be formulated as follows:
\begin{subequations}
\label{eqn:method:layer-placement}
\begin{align}
    &\sum_{i=1}^{deg} \textbf{\textit{P}}_{ui}=1,&\forall u\in \mathbb{V},\label{eqn:method:layer-placement:1}\\
    &\sum_{u\in \mathbb{V}}\textbf{\textit{P}}_{ui}\geqslant 1,&\forall i\in\{1,\dots,deg\},\label{eqn:method:layer-placement:2}\\
    &\textbf{\textit{P}}_{ui}\in\{0,1\},&\forall u\in \mathbb{V},~i\in\{1,\dots,deg\}.\label{eqn:method:layer-placement:3}
\end{align}
\end{subequations}

\noindent\textbf{Strategy-selection constraint}\quad 
Each layer must and can choose only one strategy. This constraint can be formulated as follows:\begin{subequations}
\label{eqn:method:strategy-selection}
\begin{align}
    &\sum_{k = 1}^{\lvert \mathbb{S}_u\rvert}\textbf{\textit{S}}_{uk}=1,&\forall u\in \mathbb{V},\label{eqn:method:strategy-selection:1}\\
    &\textbf{\textit{S}}_{uk}\in\{0,1\},&\forall u\in \mathbb{V},~k\in \{1,\dots,|\mathbb{S}_u|\}.\label{eqn:method:strategy-selection:2}
\end{align}
\end{subequations}
The MIQP formulation for UniAP includes the objective function in Equation~\eqref{eqn:method:lp-contig} and all the constraints from Euqation~\eqref{eqn:method:intra-stage}~-~\eqref{eqn:method:strategy-selection:2}.
 %UniAP eventually obtains the minimum TPI and its corresponding parallel strategies by solving the MIQP expression using an off-the-shelf solver. For a visualization of a potential solution to MIQP, please refer to Appendix \ref{appendix:visualization_for_p_and_s}.
\begin{algorithm}[t]
    \caption{Unified Optimization Process}
    \label{alg:unified-opt-proc}
\begin{algorithmic}
    \STATE {\bfseries Input:} Profiling results $PR$, strategy dictionary $SD$, mini-batch size $B$, computation graph $\mathcal{G}$, and the number of GPUs $n$.
    \STATE {\bfseries Output:} Optimal cost $cost^{*}$, pipeline parallel size $deg^{*}$, the number of micro-batches $c^{*}$, layer placement $\textbf{\textit{P}}^{*}$, and intra-layer strategy $\textbf{\textit{S}}^{*}$.
    \STATE $deg^{*} = 1$;
    \STATE $c^{*} = B$;
    \STATE $A$, $R$, \_, $M$ = \verb+CostModeling+($PR$, $SD[1]$, $\mathcal{G}$, $B$);
    \STATE $cost^{*}$, $P^{*}$, $S^{*}$ = \verb+QIP+($\textbf{\textit{A}}$, $\textbf{\textit{R}}$, $\textbf{\textit{M}}$);
    \STATE Get all factors for $n$ except 1 and insert them to set $\mathbb{F}$;
    \STATE Get all factors for $B$ except 1 and insert them to set $\mathbb{B}$;
    \FOR{$deg$ \textbf{in} $\mathbb{F}$}
        \FOR{$c$ \textbf{in} $\mathbb{B}$}
            \STATE Micro-batch size $b = B/c$;
            \STATE $\textbf{\textit{A}}$, $\textbf{\textit{R}}$, $\textbf{\textit{R}}'$, $\textbf{\textit{M}}$ = \verb+CostModeling+($PR$, $SD[deg]$, $\mathcal{G}$, $b$);
            \STATE $cost$, $\textbf{\textit{P}}$, $\textbf{\textit{S}}$ = \verb+MIQP+($\textbf{\textit{A}}$, $\textbf{\textit{R}}$, $\textbf{\textit{R}}'$, $\textbf{\textit{M}}$, $deg$, $c$);
            \IF{$cost < cost^{*}$}
                \STATE $cost^{*}$, $deg^{*}$, $c^{*}$, $\textbf{\textit{P}}^{*}$, $\textbf{\textit{S}}^{*}$ = $cost$, $deg$, $c$, $\textbf{\textit{P}}$, $\textbf{\textit{S}}$;
            \ENDIF
        \ENDFOR
    \ENDFOR
\end{algorithmic}
\end{algorithm}

\subsection{Unified optimization process}\label{subsec:method:unified-optimization}
UOP integrates the cost model and MIQP based on the profiling results and the computation graph to return the optimal parallel strategy and the corresponding TPI. Algorithm \ref{alg:unified-opt-proc} summarizes the whole process. In Algorithm \ref{alg:unified-opt-proc}, we denote intra-layer cost as $\textbf{\textit{A}}$, inter-layer cost as $\textbf{\textit{R}}$, cross-stage cost as $\textbf{\textit{R}}'$, and memory cost as $\textbf{\textit{M}}$. The \verb+CostModeling+ process calculates these four costs based on the cost model described in Section~\ref{subsec:method:cost-model}. 

First, UOP optimizes intra-layer-only parallelism for cases in which pipeline parallelism is not adopted. Several works~\citep{zheng_alpa_2022,liu_colossal-auto_2023} have used quadratic integer programming~(QIP) to optimize intra-layer-only parallel strategy and achieved promising results. UniAP provides a QIP formulation for intra-layer-only parallelism in Appendix \ref{appendix:miqp-for-intra-layer-parallelism}.

Then, UOP enumerates all factors of $n$ except 1 as the pipeline parallel size $deg$. For each $deg$, UOP enumerates all factors of $B$ except 1 as the number of micro-batches $c$. These enumerations aim to achieve load balance on a homogeneous cluster.

For each candidate $deg$ and $c$, UOP formulates the cost for a training iteration to an MIQP expression. It then waits for the MIQP solver to return the optimal cost and parallel strategy. 

Finally, UOP returns the minimum cost $cost^{*}$ and its corresponding pipeline parallel size $deg^{*}$, number of micro-batches $c^{*}$, layer placement $\textbf{\textit{P}}^{*}$, and intra-layer strategies $\textbf{\textit{S}}^{*}$. We provide visualization for a candidate solution to UOP in Appendix~\ref{appendix:visualization_for_p_and_s}.

\subsection{Complexity analysis}\label{subsec:method:complexity-analysis}
Let $\lvert \mathbb{V}\rvert$, $\lvert \mathbb{S}\rvert$, and $n$ denote the number of layers, parallel strategies, and GPUs, respectively. As illustrated in Algorithm \ref{alg:unified-opt-proc}, UniAP enumerates all factors of $n$ except 1 as the $deg$ in the outer loop and all factors of $B$ except 1 in the inner loop. The time complexity for the two loops in the UniAP algorithm is $\mathcal{O}(\sqrt{Bn})$. Within the inner loop body, UniAP calls \verb+CostModeling+ to model the cost of each stage for each parallel strategy. Furthermore, the optimization time limit of the MIQP solver can be set as a constant hyperparameter when UniAP calls it. Therefore, the overall computational complexity of UniAP is $\mathcal{O}(\lvert \mathbb{V} \rvert \lvert \mathbb{S}\rvert \sqrt{Bn})$.

%% file: sec/4_experiments.tex
\section{Experiment}
\label{sec:experiment}
\subsection{Experiment setup}
\label{sec:experiment-setup}
We conduct experiments on four different kinds of environments to validate the effectiveness and universality of our method. \textsc{EnvA} refers to a node~(machine) with 1 Xeon 6248 CPU, 8 V100-SXM2 32GB GPUs, and 472GB memory. \textsc{EnvB} refers to 4 nodes interconnected with 10Gbps networks and each node has 2 Xeon E5-2620 v4 CPUs, 4 TITAN Xp 12GB GPUs, and 125GB memory. \textsc{EnvC} refers to a node with 8 A100 40GB PCIe GPUs.  \textsc{EnvD} is a cloud cluster with 16 nodes and each node has 4 Hygon DCUs~(a type of non-NVIDIA GPU). In the following text, we specify the number of GPUs after the environment name (e.g., \textsc{EnvB-8}) only when partial environment nodes are used. 

We evaluate UniAP with five Transformer-based models, BERT-Huge~\citep{devlin_bert_2019}, T5-Large~\citep{raffel_exploring_2020}, ViT-Huge~\citep{dosovitskiy_image_2021}, Swin-Huge~\citep{liu_swin_2021}, and Llama~\citep{touvron_llama_2023,touvron_llama_2023-1}. We follow the common practice of training these transformer-based models. To eliminate factors that affect training throughput, we turn off techniques orthogonal to parallel strategies, such as activation checkpointing~\citep{chen_training_2016}. However, we integrate FP16 mixed precision training~\citep{micikevicius_mixed_2018} for the largest model, Llama, in some cases to successfully orchestrate the process. More details on these models are provided in Appendix~\ref{appendix:experiment-settings}.

Several solvers can be used for QIP and MIQP optimization. In our implementation, we choose Gurobi~\citep{gurobi_optimization_llc_gurobi_2023}, with detailed configurations provided in Appendix~\ref{appendix:experiment-settings}. Exploring other solvers is left for future work.
%Our implementation adopts the Gurobi Optimizer~\citep{gurobi_optimization_llc_gurobi_2023} as our QIP and MIQP solver. The configurations for Gurobi can be found in Appendix~\ref{appendix:experiment-settings}. There are some alternative solvers, such as CBC and GLPK. We leave exploring these solvers for future work.

The experimental evaluation concentrates on two primary metrics: training throughput and strategy optimization time. The former is calculated by averaging throughput from the 10th to the 60th iteration of training, while the latter is determined by measuring the time of the UOP. More details are provided in Appendix \ref{appendix:experiment-settings}.  Besides, we provide the accuracy of our cost model in Appendix \ref{appendix:experiment:performance_modeling}.

\begin{table*}[t]
    \begin{center}
    \begin{small}
    \begin{threeparttable}
    \begin{tabular}{cccccccc}
    \toprule
     \multirow{2}[0]{*}{Env.} & \multirow{2}[0]{*}{Model} & \multicolumn{3}{c}{Training throughput (samples/s)} & \multirow{2}[0]{*}{\makecell{Minimum\\speedup}} & \multirow{2}[0]{*}{\makecell{Maximum\\speedup}}\\
     \cmidrule(r){3-5}
    & & Galvatron & Alpa & UniAP &  \\
    \midrule
   \multirow{4}[0]{*}{\textsc{EnvA}} & BERT-Huge &  \textbf{33.46~$\pm$~0.28 } &  31.56~$\pm$~0.04 &  \textbf{33.46~$\pm$~0.28 } & 1.00 &  1.06 \\
    & T5-Large & \textbf{23.29~$\pm$~0.04} &  \texttt{MEM}$\times$\tnote{2)} &  \textbf{23.29~$\pm$~0.04} & 1.00 &  1.00 \\
    & ViT-Huge &   \textbf{109.51~$\pm$~0.07 } &  97.66~$\pm$~1.42 &  \textbf{109.51~$\pm$~0.07 } & 1.00 &  1.12 \\
    & Swin-Huge & \verb+CUDA+$\times$\tnote{3)} & \verb+N/A+\tnote{4)}  &  \textbf{67.96~$\pm$~0.12 } & \verb+N/A+\tnote{4)} & \verb+N/A+\tnote{4)}\\
    \midrule
    \multirow{4}[0]{*}{\textsc{EnvB-8}}
    & BERT-Huge &  6.27~$\pm$~0.17 &  8.95~$\pm$~0.06 &  \textbf{10.77~$\pm$~0.13} &  1.20 &  1.71 \\
    & T5-Large\tnote{1)} &  \textbf{8.06~$\pm$~0.06} & \verb+MEM+$\times$\tnote{2)}  &  7.98~$\pm$~0.05 &  0.99  &  0.99\\
    & ViT-Huge  &  32.20~$\pm$~0.17 &  38.74~$\pm$~0.20 &  \textbf{45.58~$\pm$~0.54} &  1.18 &  1.41 \\
    & Swin-Huge &  13.90~$\pm$~0.17 & \verb+N/A+\tnote{4)} &  \textbf{19.08~$\pm$~0.10} &  1.37 &  1.37\\
    \midrule
    \textsc{EnvC} & Llama-7B & 1.22~$\pm$~0.01 & \verb+N/A+\tnote{4)} & \textbf{4.63~$\pm$~0.007} & 3.80 & 3.80 \\
    \midrule
    \midrule
    \multirow{2}[0]{*}{Env.} & \multirow{2}[0]{*}{Model} & \multicolumn{3}{c}{Strategy optimization time (min.)} & \multirow{2}[0]{*}{\makecell{Minimum\\speedup}} & \multirow{2}[0]{*}{\makecell{Maximum\\ speedup}}\\
     \cmidrule(r){3-5}
    & & Galvatron & Alpa & UniAP &  \\
    \midrule
    \multirow{4}[0]{*}{\textsc{EnvA}}
    & BERT-Huge &  6.44~$\pm$~0.588 &  $>$~40 &  \textbf{0.37~$\pm$~0.002} &  17.29 &  $>$~107.41\\
    & T5-Large &  12.41~$\pm$~0.122 &  \texttt{MEM}$\times$\tnote{2)} &  \textbf{0.89~$\pm$~0.007} &  13.98 &  13.98\\
    & ViT-Huge &  6.29~$\pm$~0.464 &  $>$~40 &  \textbf{0.57~$\pm$~0.009} &  10.95 &  $>$~69.60 \\
    & Swin-Huge &  11.88~$\pm$~0.666 & \verb+N/A+\tnote{4)} &  \textbf{2.16~$\pm$~0.004} &  5.49 &  5.49 \\
    \midrule
    \multirow{4}[0]{*}{\textsc{EnvB-8}}
    & BERT-Huge &  2.04~$\pm$~0.010 &  $>$~40 &  \textbf{1.51~$\pm$~0.005} &  1.34 &  $>$~26.32\\
    & T5-Large\tnote{1)} &  2.64~$\pm$~0.110 & \verb+MEM+$\times$\tnote{2)} &  \textbf{0.91~$\pm$~0.005} &  2.90 &  2.90\\
    & ViT-Huge &  2.37~$\pm$~0.180 &  $>$~40 &  \textbf{1.11~$\pm$~0.011} &  2.14 &  $>$~36.01\\
    & Swin-Huge &  4.29~$\pm$~0.320 & \verb+N/A+\tnote{4)} &  \textbf{2.29~$\pm$~0.010} &  1.87 &  1.87\\
    \midrule
    \textsc{EnvC} & Llama-7B & 6.84~$\pm$~0.055 & \verb+N/A+\tnote{4)} & \textbf{0.58~$\pm$~0.006} & 11.83 & 11.83 \\
    \bottomrule
    \end{tabular}
    \begin{tablenotes}
    \footnotesize
    \item[1)] T5-Large tested on \textsc{EnvB} is restricted to 16/16 layers to avoid out-of-memory~(OOM) exceptions.
    \item[2)] \texttt{MEM}$\times$: OOM exceptions during strategy optimization.
    \item[3)] \texttt{CUDA}$\times$: CUDA OOM exceptions during model training.
    \item[4)] The official implementation for Swin-Huge and Llama in Alpa is absent. We have endeavored to adopt the code, yet encountered compilation errors. Consequently, experiments related to these models on Alpa are marked as \texttt{N/A}.
    \end{tablenotes}
    \end{threeparttable}
\end{small}
\end{center}
\vskip -0.2in
\caption{Training throughput and strategy optimization time on \textsc{EnvA}, \textsc{EnvB-8}, and \textsc{EnvC}.}
\label{tab:throughput_and_optimization_time}
\end{table*}

\begin{table*}[t]
\begin{center}
\begin{small}
\begin{threeparttable}
\begin{tabular}{ccccccc}
    \toprule
    \multirow{2}[0]{*}{Model}  & \multicolumn{3}{c}{\makecell{Training throughput (samples/s)}} & \multicolumn{3}{c}{\makecell{Strategy optimization time (min.)}} \\
    \cmidrule(lr){2-4}\cmidrule(lr){5-7}
           & Megatron & DeepSpeed & UniAP & Megatron & DeepSpeed & UniAP \\
    \midrule
    Llama-7B  & \textbf{2.01~$\pm$~0.005}  & \texttt{SOL}$\times$\tnote{1)}  & \textbf{2.01~$\pm$~0.005}  & > 8.0 hours & \texttt{SOL}$\times$\tnote{1)}  & \textbf{3.07~$\pm$~0.121}  \\
    Llama-13B &  \textbf{0.82~$\pm$~0.001}  &   \texttt{SOL}$\times$\tnote{1)}    &  \textbf{0.82~$\pm$~0.001}     &    > 2.5 hours   &    \texttt{SOL}$\times$\tnote{1)}   & \textbf{1.95~$\pm$~0.076} \\
\bottomrule
\end{tabular}

\begin{tablenotes}
    \footnotesize
    \item[1)] \texttt{SOL}$\times$: No solution after strategy optimization.

\end{tablenotes}

\end{threeparttable}
\end{small}
\end{center}
\vskip -0.2in
\caption{Training throughput and strategy optimization time on \textsc{EnvD-32}.}
\label{tab:dcu}%
\vskip -0.2in
\end{table*}

% \begin{figure*}[h]
% \centering
% \begin{minipage}[t]{0.48\linewidth}
% \centering
% \includegraphics[width=0.9\linewidth]{performance_modeling.pdf}
% \caption{Relative estimation error.}
% \label{fig:performance-modeling}
% \end{minipage}
% \hskip 0.1in
% \begin{minipage}[t]{0.48\linewidth}
% \centering
% \includegraphics[width=\columnwidth]{ablation.pdf}
% \caption{Ablation study. \texttt{SOL}$\times$ indicates no feasible solution during strategy optimization, and \texttt{CUDA}$\times$ indicates CUDA OOM exceptions during training.}
% \label{fig:ablation}
% \end{minipage}
% \end{figure*}

\begin{table*}[t]
\begin{center}
\begin{small}
\begin{threeparttable}
\begin{tabular}{p{2.5em}cp{2.5em}ccccccc}
    \toprule
    &\multirow{2}[0]{*}{Model}& & \multirow{2}[0]{*}{Batch size} & \multicolumn{4}{c}{Training throughput (samples/s)} & \multirow{2}[0]{*}{\#infeasible\tnote{5)}} & \multirow{2}[0]{*}{\#candidate\tnote{6)}} \\
     \cmidrule(lr){5-8}
    & && & 1 st.\tnote{1)} & 2 nd.\tnote{2)} & Slowest\tnote{3)} & Median\tnote{4)} & & \\
    \midrule
    &Llama-7B& & 8 & 2.01 & 1.92 & 0.22  & 0.82  & 41 & 64 \\
    &Llama-13B& & 4 & 0.82 & 0.58  & 0.27  & 0.42  &  42  & 48 \\
\bottomrule
\end{tabular}
\begin{tablenotes}
    \footnotesize
    \item[1)] 1 st.: The training throughput achieved by the fastest parallel strategy.
    \item[2)] 2 nd.: The training throughput achieved by the second fastest parallel strategy.
    \item[3)] Slowest: The training throughput achieved by the slowest parallel strategy.
    \item[4)] Median: The median value of training throughputs across parallel strategies that will successfully train the model.
    \item[5)] \#infeasible: The number of parallel strategies that will encounter exceptions such as CUDA OOM during model training.
    \item[6)] \#candidate: The total number of candidate parallel strategies.
\end{tablenotes}
\end{threeparttable}
\end{small}
\end{center}
\vskip -0.2in
\caption{Statistics on the candidate parallel strategies for Megatron.}
\label{tab:statistics-on-megatron}%
\vskip -1.5em
\end{table*}

\subsection{Training throughput and strategy optimization time}\label{subsec:experiment:optimization-time-and-throughput}
We compare the training throughput and strategy optimization time of UniAP with those of the baselines on \textsc{EnvA}, \textsc{EnvB-8}, \textsc{EnvC}, and \textsc{EnvD-32}. For experiments on \textsc{EnvA}, we set the mini-batch size to be 32, 16, 128, and 128 for BERT, T5, ViT, and Swin, respectively. For experiments on \textsc{EnvB-8}, we set the mini-batch size to be 16, 8, 64, and 32 for these four models, respectively. For the Llama model~\citep{touvron_llama_2023,touvron_llama_2023-1} run on \textsc{EnvC}, we set the mini-batch size to be 8. For experiments on \textsc{EnvA}, \textsc{EnvB-8}, and \textsc{EnvC}, we choose Galvatron~\citep{miao_galvatron_2022} and Alpa~\citep{zheng_alpa_2022} as baselines because they have achieved SOTA performance. Specifically, Galvatron has surpassed other methods, including PyTorch DDP~\citep{li_pytorch_2020}, Megatron-LM~\citep{narayanan_efficient_2021}, FSDP~\citep{rajbhandari_zero_2020,fairscale_authors_fairscale_2021}, GPipe~\citep{huang_gpipe_2019}, and DeepSpeed 3D~\citep{deepspeed-3d} in terms of training throughput, as reported in the original paper~\citep{miao_galvatron_2022}. Furthermore, Alpa utilizes the Just-In-Time~(JIT) compilation feature in JAX and outperforms Megatron-LM and DeepSpeed.

For benchmarks on  \textsc{EnvD-32}, pytorch libraries and frameworks need specialized adaptations due to the heterogeneity of DCUs against NVIDIA GPUs. Thus, this part of experiments is based on adapted pytorch libraries and Megatron-DeepSpeed\citep{deepspeed-3d} framework provided by the platform developers, and we select MP methods including Megatron~\citep{narayanan_efficient_2021} and DeepSpeed~(ZeRO-3)~\citep{rasley_deepspeed_2020} as our baseline methods. We set the mini-batch size as 8 for Llama-7B and 4 for Llama-13B. The remaining configurations align with those for Llama-7B in Section~\ref{sec:experiment-setup}, such as activating FP16 mixed precision training.
 
Table~\ref{tab:throughput_and_optimization_time} shows the training throughput and strategy optimization time on \textsc{EnvA}, \textsc{EnvB-8}, and \textsc{EnvC}. On \textsc{EnvA}, UniAP and Galvatron get the same optimal strategy for BERT-Huge, T5-Large, and ViT-Huge, outperforming Alpa in terms of training throughput and strategy optimization time. In addition, UniAP finds a solution for Swin-Huge, while Galvatron encounters CUDA OOM issues. In particular, UniAP achieves a maximum optimization speedup that is 17$\times$ faster than Galvatron and hundreds of times faster than Alpa on BERT-Huge. This is mainly due to the ability of the MIQP solver to search for an optimal strategy on multiple threads, while the dynamic programming based methods like Galvatron and Alpa run on a single thread due to their strong data dependency.

On \textsc{EnvB-8}, UniAP consistently demonstrates competitive or larger training throughput compared to Galvatron and Alpa. We attribute the performance improvement to UniAP's larger strategy space. A detailed study is provided in Section~\ref{subsec:experiment:further study}. Furthermore, UniAP's strategy optimization time is also significantly shorter than the two baseline methods.

On \textsc{EnvC}, UniAP shows an optimization speedup of 11.83$\times$ and a training speedup of 3.80$\times$ compared to Galvatron on Llama. We examine the parallel strategy they adopted. UniAP employs an 8-stage PP with a micro-batch size of 1, whereas Galvatron employs a 4-stage PP with a micro-batch size of 3 (with the final micro-batch containing 2 samples). Within each PP stage, Galvatron utilizes a 2-way TP. For \textsc{EnvC}, which is equipped with 8 A100 40GB PCIe GPUs, minimizing communication volume is critical. Since PP has significantly less inter-device communication volume than TP, the parallel strategy discovered by UniAP is more reasonable than Galvatron in this case.

Table~\ref{tab:dcu} shows the results on \textsc{EnvD-32}. On this environment, UniAP consistently identifies the fastest parallel strategies while expending considerably less time on strategy optimization compared to Megatron~(about 157$\times$ faster for Llama-7B and 77$\times$ faster for Llama-13B). Please note that the strategy optimization time Megatron needs for Llama-7B surpasses that for Llama-13B. We attribute this phenomenon to the variations in the mini-batch size. Specifically, the escalation from a mini-batch size of 4 (employed in Llama-13B) to 8 (employed in Llama-7B) leads to a rise in the number of candidate parallel strategies, as well as the parallel strategies that will successfully train the model. As a result, the strategy optimization time will become increasingly long when the mini-batch size exceeds 8 in the pretraining scenario~\citep{brown_language_2020,touvron_llama_2023,touvron_llama_2023-1}. In this situation, UniAP will identify the optimal parallel strategy much faster.

Furthermore, our experiments highlight a limitation encountered with DeepSpeed~(ZeRO-3). It requires the mini-batch size to be divisible evenly by the total number of computation devices. This specific prerequisite prevents DeepSpeed from successfully launching the training process with 32 DCUs.

To facilitate further discussions, we provide a case study of the optimal parallel strategy in Appendix~\ref{appendix:case-study}.

\begin{figure}[t]
\begin{center}
\begin{subfigure}{0.49\linewidth}
\centering
\includegraphics[width=\linewidth]{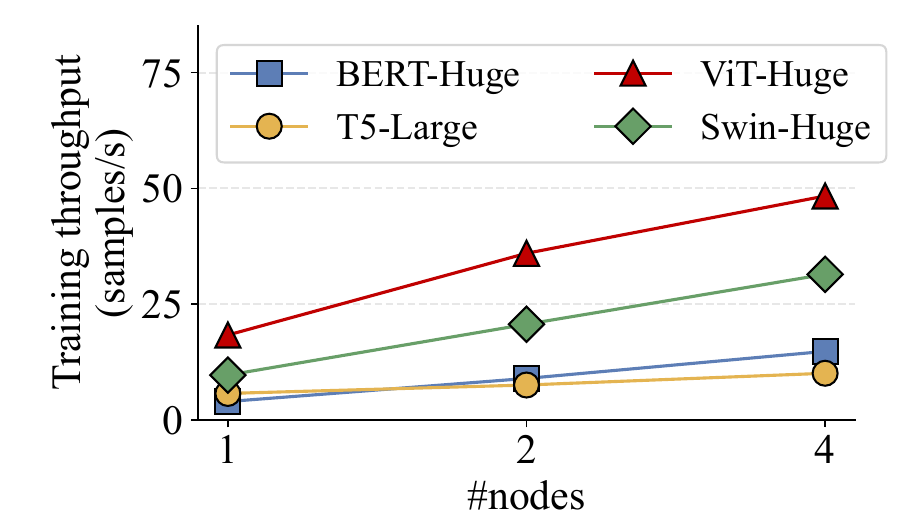}
\caption{Training throughput.}
\label{fig:scalability:throughput}
\end{subfigure}
\begin{subfigure}{0.49\linewidth}
\centering
\includegraphics[width=\linewidth]{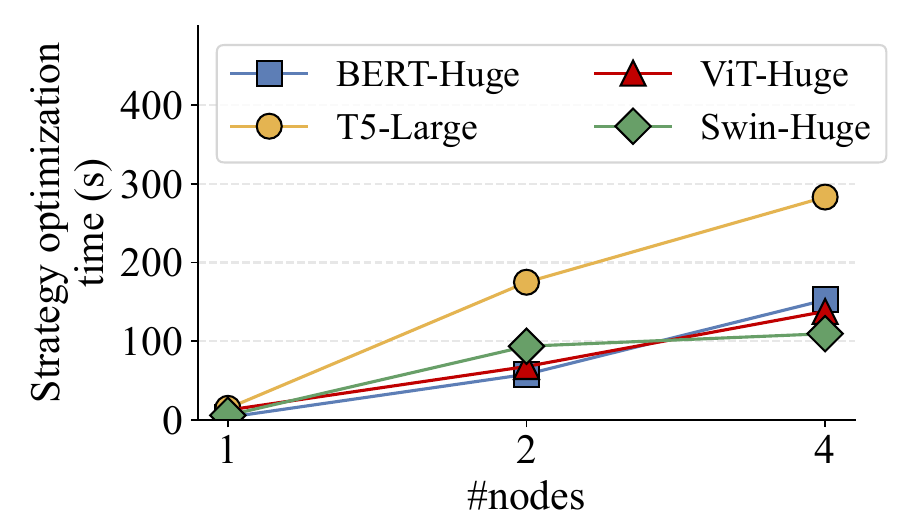}
\caption{Strategy optimization time.}
\label{fig:scalability:optimization-time}
\end{subfigure}
\vskip -0.1in
\caption{Scalability on training throughput and strategy optimization time.}
\vskip -0.1in
\label{fig:scalability}
\end{center}
\end{figure}

\begin{table}[t]
\begin{center}
\begin{small}
\begin{threeparttable}
\begin{tabular}{ccc}
    \toprule
    Number of & Training throughput  & Strategy optimization\\
    DCUs &(samples/s) & time(min.) \\
    \midrule
    16 &0.1726~$\pm$~0.0005&0.75~$\pm$~0.0100 \\
    32 &0.3393~$\pm$~0.0004&1.64~$\pm$~0.0321 \\
    64 &0.6656~$\pm$~0.0023&1.55~$\pm$~0.0058 \\
\bottomrule
\end{tabular}
\end{threeparttable}
\end{small}
\end{center}
\vskip -0.15in
\caption{Scalability on training throughput among DCU clusters.}
\label{tab:scalability-on-dcu}%
\vskip -1em
\end{table}

\subsection{Scalability}\label{subsec:experiment:scalability}

We study the scalability of UniAP using BERT, T5, ViT, and Swin on \textsc{EnvB}. We set the mini-batch sizes for each data point as 8, 4, 32, and 16 times the number of nodes~(denoted as `\#nodes'). The experimental results are shown in Figure~\ref{fig:scalability}. In Figure~\ref{fig:scalability:throughput}, the training throughput of the optimal strategy demonstrate near-linearity as the number of nodes and mini-batch size increase. In Figure~\ref{fig:scalability:optimization-time}, the strategy optimization time matches the complexity analysis in Section~\ref{subsec:method:complexity-analysis}.

Additionally, we use DCU nodes in \textsc{EnvD} to scale UniAP to larger clusters on Llama-7B, with precision of FP32. We conducted experiments on clusters with 16, 32, and 64 DCUs, where we set the mini-batch size to 1/8 of the number of DCUs (i.e., for a cluster with 16 DCUs, we set the mini-batch size to 2). The experimental results are shown in Table \ref{tab:scalability-on-dcu}, further demonstrating the near-linear scalability of UniAP.

% and the corresponding strategy optimization time demonstrate near-linearity This phenomenon reveals that UniAP is scalable and verifies the computational complexity analysis in Section~\ref{subsec:method:unified-optimization}.

\subsection{Further study on the significance of UniAP}
\label{subsec:experiment:further study}
Finally, we investigate the significance of UniAP by answering two questions.
The first question is why we need AP rather than MP. We examine the statistics on the candidate parallel strategies for Megatron, as shown in Table~\ref{tab:statistics-on-megatron}. With MP, most users are unfamiliar with the complex hyper-parameter detail of the candidate parallel strategies and typically choose a parallel strategy at random, which will result in a 64.1\% (41/64) chance of failing to train Llama-7B and an 87.5\% (42/48) chance of failing to train Llama-13B even when the hardware resources are sufficient. Even if they successfully eliminate all infeasible parallel strategies, they still only have a 50\% chance of identifying a parallel strategy with training throughput better than 0.82 samples/s for Llama-7B and 0.42 samples/s for Llama-13B. When the users can't identify the fastest parallel strategy out of hundreds of candidates, they may sacrifice at least 4.4\% of the training throughput for Llama-7B and 29.2\% for Llama-13B. In such circumstances, UniAP is able to identify the fastest strategy, as shown in Table~\ref{tab:dcu}. 

\begin{figure}[t]
\centering
\includegraphics[width=\columnwidth]{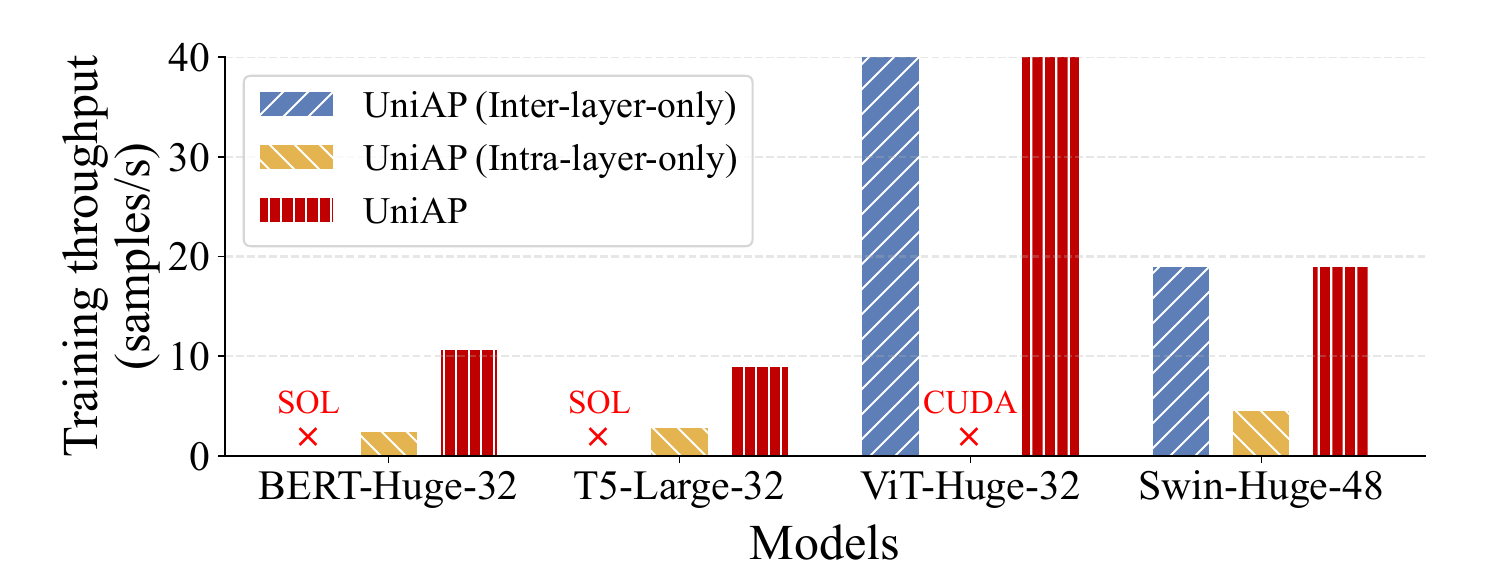}
\caption{Ablation study. \texttt{SOL}$\times$ indicates no feasible solution during strategy optimization, and \texttt{CUDA}$\times$ indicates CUDA OOM exceptions during training.}
\label{fig:ablation}
\vskip -1.5em
\end{figure}

%\begin{table}
%\vskip -0.2in
%\caption{Ablation study on the importance of unifying strategy space.}
%\label{tab:ablation}%
%\begin{center}
%\begin{small}
%\begin{threeparttable}
%    \begin{tabular}{cccc}
%    \toprule
%    \multirow{3}[0]{*}{Model} & \multicolumn{3}{c}{Training throughput (samples/s)} \\
%    \cmidrule(r){2-4}
%          & UniAP (Inter-only) & UniAP (Intra-only) & UniAP \\
%    \midrule
%    BERT-Huge &  \texttt{SOL}$\times$\tnote{1)} &  2.48~$\pm$~0.02 &  \textbf{10.77~$\pm$~0.13} \\
%    T5-Large &  \texttt{SOL}$\times$\tnote{1}   &  2.92~$\pm$~0.01 &  \textbf{9.01~$\pm$~0.06} \\
%    ViT-Huge &  \textbf{45.58~$\pm$~0.54} & \verb+CUDA+$\times$\tnote{2}   &  \textbf{45.58~$\pm$~0.54} \\
%    Swin-Huge &  \textbf{19.08~$\pm$~0.10} &  4.66~$\pm$~0.02 &  \textbf{19.08~$\pm$~0.10} \\
%    \bottomrule
%\end{tabular}%
%\begin{tablenotes}
    %\footnotesize
    %\item[1)] \texttt{SOL}$\times$: No solution after strategy optimization.
    %\item[2)] \texttt{CUDA}$\times$: CUDA OOM exceptions during model training.
%\end{tablenotes}
%\end{threeparttable}
%\end{small}
%\end{center}
%\vskip -0.2in
%\end{table}%

The second question is why UniAP achieves better performance than other AP methods.
We investigate the importance of the strategy space for the optimality of parallel strategies with an ablation study. Specifically, we constrain the strategy space to inter-layer-only and intra-layer-only strategies and evaluate the training throughput of the resulting optimal strategy on \textsc{EnvB}. We set the mini-batch sizes to be 16, 12, 64, and 32, respectively. 
Results are shown in Figure~\ref{fig:ablation}. We can find that constraining the strategy space compromises the optimality of parallel strategies or gets strategies that encounter OOM across different models. 
%For example, constraining the strategy space to inter-layer-only parallelism yield infeasible solution for BERT-Huge on \textsc{EnvD}. Meanwhile, constraining the strategy space to intra-layer-only parallelism may reduce the training throughput by 4.34$\times$ for BERT-Huge on \textsc{EnvD}.
Hence, unifying inter- and intra-layer AP for joint optimization makes UniAP outperform other AP methods. 

%% file: sec/5_conclusion.tex
\section{Conclusion}\label{sec:conclusion}
In this paper, we propose a novel AP method called UniAP to unify inter- and intra-layer AP by MIQP.
To the best of our knowledge, UniAP is the first parallel method that can jointly optimize the two categories of parallel strategies to find an optimal solution.
Experimental results show that UniAP can outperform other state-of-the-art baselines to achieve the best performance.

%% file: sec/7_acknowledgement.tex
\section{Acknowledgement}\label{sec:acknowledgment}
This work is supported by National Key R\&D Program of China (No.\hspace{0pt}2020YFA0713900), NSFC Project (No.\hspace{0pt}12326615), the Key Major Project of the Pengcheng Laboratory (No.\hspace{0pt}PCL2024A06).

%% file: sec/X_suppl.tex
\newpage
\appendix
\onecolumn

{\fontsize{20pt}{30pt}\selectfont
Appendix
}

\section{Notation table}\label{appendix:notation-table}
Table~\ref{tab:notation-table} summarizes the main notations used throughout the paper.

\begin{table}[h]
\begin{center}
\begin{small}
\begin{threeparttable}
\begin{tabular}{cc}
    \toprule
    Symbol & Description\\
    \midrule
    $m_a$ &the activation memory cost \\
    $m_{s}$ &the memory cost of model states \\
    $ps$ &parameter size \\
    $ts$ &TP size \\
    $fs$ &FSDP size \\
    $c_{dtype}$ &a constant dependent on mixed precision choice \\
    $m_c$ &context memory cost \\
    $deg$ &the number of computation stages \\
    $c$ &the number of micro-batches \\
    $tpi_{gpipe}$ &TPI in GPipe \\
    $m$ &the memory limit for each device \\
    $B$ &mini-batch size \\
    $n$ &the number of GPUs \\
    $fp_i$ &forward computation time for computation stage $i$ \\
    $bp_i$ &backward computation time for computation stage $i$ \\
    $fo_j$ &forward communication time for communication stage $j$ \\
    $bo_j$ &backward communication time for communication stage $j$ \\
    $p_i$ &the cost for computation stage $i$ \\
    $o_j$ &the cost for communication stage $j$ \\
    $\textbf{\textit{M}}_{uk}$ &the memory cost for the $k$-th intra-layer strategy of layer $u$ on a single device \\
    $\textbf{\textit{A}}_{uk}$ &execution cost for the $k$-th intra-layer strategy of layer $u$ \\
    $\textbf{\textit{S}}_{uk}$ &whether the $k$-th parallel strategy is selected for the layer $u$ \\
    $\textbf{\textit{P}}_{ui}$ &whether layer $u$ is placed on the $i$-th computation stage \\
    $\textbf{\textit{R}}_{uv}$ &resharding cost between layers $u$ and $v$ when they are located within the same pipeline stage \\
    $\textbf{\textit{R}}'_{uv}$ &resharding cost between layers $u$ and $v$ when they are located across consecutive stages \\
    $\textbf{\textit{Z}}$ &the auxiliary variable used for order-preserving constraint \\
    $\mathbb{P}$ &the set of cost for computation stages, which is $\{p_1, p_2, \dots, p_{deg}\}$ \\
    $\mathbb{O}$ &the set of cost for communication stages, which is $\{o_1, o_2, \dots, o_{deg-1}\}$ \\
    $\mathbb{S}_u$ &the set of intra-layer parallel strategies for layer $u$ \\
    $\mathcal{G}(\mathbb{V}, \mathbb{E})$ &the computation graph for the model \\
\bottomrule
\end{tabular}
\end{threeparttable}
\end{small}
\end{center}
\vskip -0.15in
\caption{Summary of the main notations.}
\label{tab:notation-table}%
\vskip -1em
\end{table}

\section{Proof of the linear form for the contiguous set}\label{appendix:linear-form-of-contigous-constraint}
To facilitate our discussion, we adopt the linear form of the order-preserving constraint as presented in the main paper. We denote $\textbf{\textit{P}}_{ui}$ as a 0-1 variable indicating whether layer $u$ is to be placed on the $i$-th computation stage, $pp\_size$ as the number of computation stages in the pipeline. Besides, $\mathcal{G}(\mathbb{V}, \mathbb{E})$ represents the computation graph for the model. Then, we formalize the theorem as follows:

\begin{theorem}
    A subgraph with node set $\mathbb{V}_i=\{\forall u\in \mathbb{V}: \textbf{\textit{P}}_{ui}=1\}$ is contiguous if and only if there exists $\textbf{\textit{Z}}_{vi}$ such that Equation~\eqref{eqn:method:order-preserving:1}, \eqref{eqn:method:order-preserving:2}, and \eqref{eqn:method:order-preserving:3} are satisfied.
\end{theorem}

Previous work~\citep{tarnawski_efficient_2020} has proven this theorem. Our proof draws on the process of this work. The details of the proof are as follows:

\begin{proof}
\vskip -0.1in
    "If": Assume that there exists nodes $u, w\in \mathbb{V}_i$ and $v \notin \mathbb{V}_i$ such that $v$ and $w$ are reachable from $u$ and $v$, respectively. Hence, $\textbf{\textit{P}}_{ui} = 1$, $\textbf{\textit{P}}_{wi} = 1$, and $\textbf{\textit{P}}_{vi} = 0$. Without losing generality, we assume $\langle u, v\rangle\in \mathbb{E}$. Thus, according to Equation \eqref{eqn:method:order-preserving:3}, we have $\textbf{\textit{Z}}_{vi}\leqslant \textbf{\textit{P}}_{vi}-\textbf{\textit{P}}_{ui}+1=0$. By applying Equation~\eqref{eqn:method:order-preserving:2} repeatedly following the path from $v$ to $w$, we have $\textbf{\textit{Z}}_{wi}\leqslant \textbf{\textit{Z}}_{vi}$. Thus, $\textbf{\textit{Z}}_{wi}\leqslant 0$. However, we also have $\textbf{\textit{Z}}_{wi}\geqslant \textbf{\textit{P}}_{wi}=1$ according to Equation~\eqref{eqn:method:order-preserving:1}. A contradiction.

    "Only if": First, we define $\textbf{\textit{Z}}_{vi}=1$ if a node $w\in \mathbb{V}_i$ is reachable from $v~(v\in\mathbb{V})$. Otherwise, $\textbf{\textit{Z}}_{vi}=0$. Thus, Equation~\eqref{eqn:method:order-preserving:1} and \eqref{eqn:method:order-preserving:2} are satisfied according to this kind of definition. For Equation~\eqref{eqn:method:order-preserving:3}, if $\textbf{\textit{P}}_{vi}=1$, the constraint will hold true regardless of whether $\textbf{\textit{P}}_{ui}$ is $1$ or $0$. If $\textbf{\textit{P}}_{vi}=0$ and $\textbf{\textit{P}}_{ui}=0$, $\textbf{\textit{Z}}_{vi}\leqslant \textbf{\textit{P}}_{vi}-\textbf{\textit{P}}_{ui}+1=1$ will also hold true because $\textbf{\textit{Z}}_{vi}$ could be either $0$ or $1$. Finally, if $\textbf{\textit{P}}_{vi}=0$ and $\textbf{\textit{P}}_{ui}=1$, $\textbf{\textit{Z}}_{vi}=0$ will hold true because $\mathbb{V}_i$ is a contiguous set and we cannot find any $w\in \mathbb{V}_i$, such that $w$ is reachable from $v$.
\end{proof}

\section{QIP formulation for intra-layer-only parallelism}\label{appendix:miqp-for-intra-layer-parallelism}
Here we present the QIP formulation for intra-layer-only parallelism with explanations.

\textbf{Objective function}\quad In terms of intra-layer-only parallelism, there is only one computation stage involved. As a result, the objective function takes into account only the value of $p_1$. We hereby formalize the equation as
\begin{equation}
    \min\quad tpi_{gpipe}=p_1.\label{eqn:appendix:lp-no-pp}
\end{equation}

\textbf{Computation-stage constraint}\quad With only one computation stage in intra-layer-only parallelism, the communication-stage constraint can be omitted, and the computation and communication cost can be modeled for $p_1$. Thus, we could formalize the constraint as
\begin{equation}
\sum_{u\in \mathbb{V}}\textbf{\textit{S}}_{u}^\mathsf{T}\textbf{\textit{A}}_{u}+\sum_{\langle u,v\rangle\in \mathbb{E}}\textbf{\textit{S}}_{u}^\mathsf{T}\textbf{\textit{R}}_{uv}\textbf{\textit{S}}_{v}=p_1.~\label{eqn:appendix:computation-for-no-pp}
\end{equation}
In the equation, the first summation term for any $u\in \mathbb{V}$ represents the cost of choosing intra-layer strategies for all layers, while the second term represents the summation of resharding costs on all edges.

\textbf{Memory constraint}\quad Similar to the memory constraint in MIQP, it is necessary to ensure that the memory usage on a single device does not exceed its device memory bound $m$ in QIP. This restriction gives 
\begin{equation}
    \sum_{u\in V}\textbf{\textit{S}}_u^\mathsf{T} \textbf{\textit{M}}_u\leqslant m.~\label{eqn:appendix:memory-constraint-for-no-pp}
\end{equation}
It is worth noting that $m$ should be an identical constant across multiple devices if these devices are homogeneous. Otherwise, the value of $m$ varies.

\textbf{Strategy-selection constraint}\quad For intra-layer-only parallelism, the layer-placement constraint can be safely omitted because it is designed for PP. However, the strategy-selection constraint is necessary because each layer can only select one intra-layer strategy. Therefore, the strategy-selection constraint for QIP is identical to Equation~\eqref{eqn:method:strategy-selection:1} and \eqref{eqn:method:strategy-selection:2} for MIQP.

By combining objective function~\eqref{eqn:appendix:lp-no-pp} and constraints~\eqref{eqn:method:strategy-selection:1}, \eqref{eqn:method:strategy-selection:2}, \eqref{eqn:appendix:computation-for-no-pp}, and \eqref{eqn:appendix:memory-constraint-for-no-pp}, we have the QIP expression for optimizing the intra-layer-only AP. Like MIQP expression for optimizing the inter- and intra-layer AP, UniAP will eventually get the minimum TPI and corresponding parallel strategies by invoking the off-the-shelf solver.

\section{Visualization for the candidate solution}\label{appendix:visualization_for_p_and_s}
\begin{figure}[t]
\vskip 0.2in
\begin{center}
    \centerline{\includegraphics[width=0.6\linewidth]{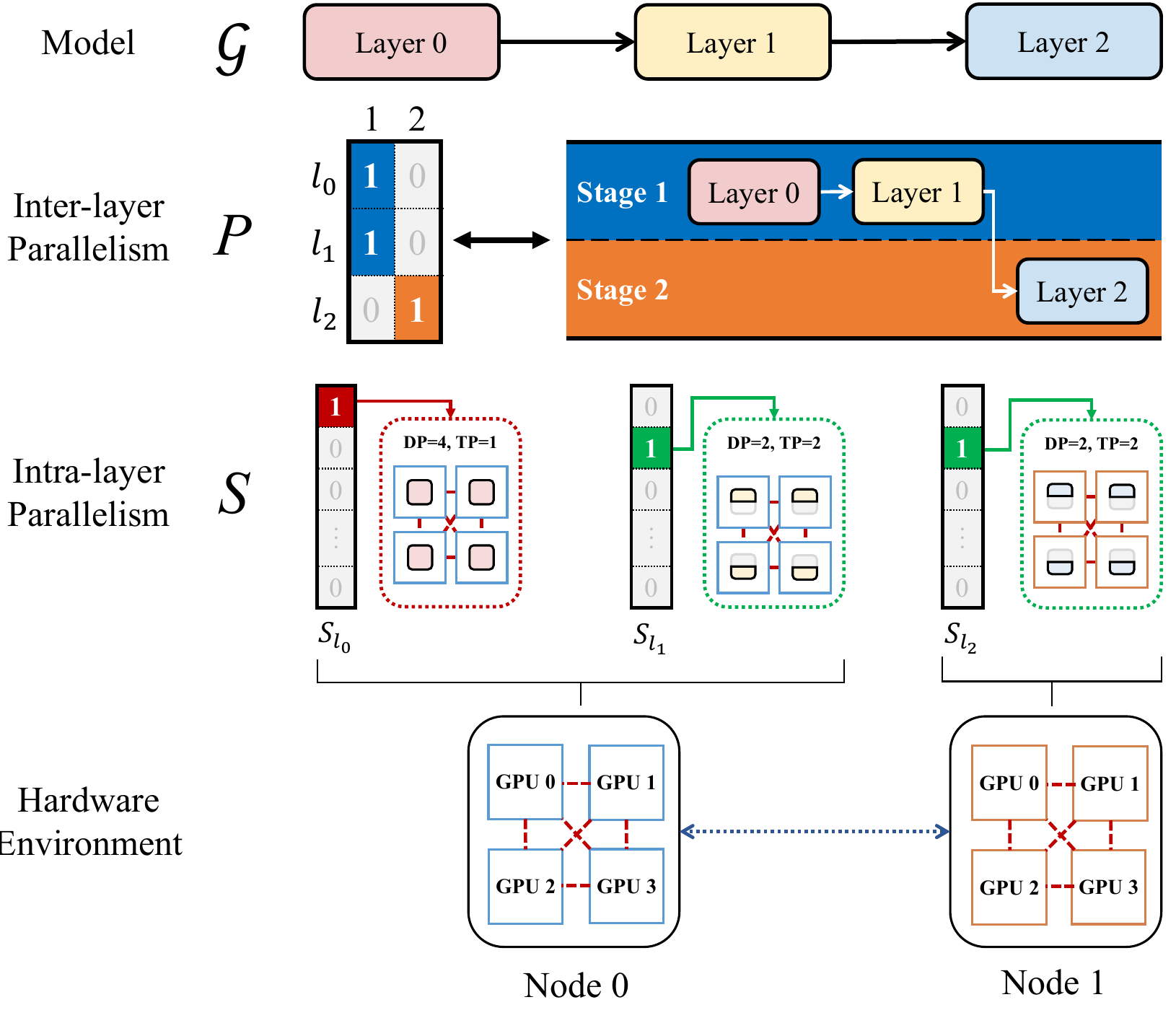}}
    \caption{A candidate solution for UOP.}
    \label{fig:ps_explanation}
\end{center}
\vskip -0.2in
\end{figure}

In this section, we proceed to visually represent a potential solution for UOP. Given a deep learning model $\mathcal{G}$, pipeline parallel size $pp\_size$, and number of micro-batches $c$, UniAP will determine layer placement $\textbf{\textit{P}}$ for inter-layer parallelism and parallel strategy $\textbf{\textit{S}}$ for intra-layer parallelism using an off-the-shelf solver. As Figure \ref{fig:ps_explanation} shows, the solver is optimizing a three-layer model with two pipeline stages, each assigned four GPUs. At this time, a candidate solution could be
\begin{equation}
    \textbf{\textit{P}}=\begin{bmatrix}
        1 & 0 \\
        1 & 0 \\
        0 & 1
    \end{bmatrix},~
    \textbf{\textit{S}}=
    \begin{bmatrix}
        1 & 0 & 0 \\
        0 & 1 & 1 \\
        0 & 0 & 0 \\ 
        \vdots & \vdots & \vdots \\
        0 & 0 & 0
    \end{bmatrix}.
\end{equation}

Here, the $u$-th row of matrix $\textbf{\textit{P}}$ denotes the placement strategy for layer $u$, where $\textbf{\textit{P}}_{ui}=1$ signifies the placement of layer $u$ on stage $i$, while $0$ indicates otherwise. For example, $\textbf{\textit{P}}_{l_{0}}=\left[1,~0\right]$ denotes the placement of layer $l_0$ on pipeline stage 1. Additionally, the $u$-th column of matrix $\textbf{\textit{S}}$ denotes the selected intra-layer parallel strategy for layer $u$, where $\textbf{\textit{S}}_{uj}=1$ denotes the selection of the $j$-th strategy from the intra-layer parallel strategy set. For example, $\textbf{\textit{S}}_{l_{0}}=\left[1,~0,~0,~\cdots,~0\right]^{\mathsf{T}}$ indicates that layer $l_0$ will adopt only the DP strategy, while $\textbf{\textit{S}}_{l_{1}}=\left[0,~1,~0,~\cdots,~0\right]^{\mathsf{T}}$ indicates that layer $l_1$ will employ a strategy where DP is performed on GPU 0, 1 and GPU 2, 3, and TP is performed across these two GPU groups.

There exist numerous combinations of $\textbf{\textit{P}}$ and $\textbf{\textit{S}}$. The off-the-shelf solver will automatically search for the optimal solution given pipeline parallel size $pp\_size$ and the number of micro-batches $c$. By solving the MIQP expression and enumerating every possible $pp\_size$ and $c$ in the UOP process, UniAP will ultimately derive an optimal parallel strategy for the deep learning model within the current hardware environment.

\section{Experiment detail}\label{appendix:experiment-settings}
\textbf{Gurobi configuration}\quad When tackling the MIQP problem, UniAP employs several configurations for the Gurobi Optimizer 10.1~\citep{gurobi_optimization_llc_gurobi_2023}. In particular, we set \textit{TimeLimit} to 60 seconds, \textit{MIPFocus} to 1, \textit{NumericFocus} to 1, and remain other configurations to default. For instance, we establish the \textit{MIPGap} parameter as the default value of 1e-4 to serve as a strict termination criterion. Furthermore, we have implemented an early stopping mechanism to terminate the optimization process as early as possible. There are two conditions that can activate the mechanism. Firstly, if the current runtime exceeds 15 seconds and the relative MIP optimality gap is less than 4\%, we will terminate the optimization. Secondly, if the current runtime exceeds 5 seconds and the best objective bound is worse than the optimal solution obtained in the previous optimization process, we will terminate the optimization.

% \begin{table}[t]
% \caption{Details for five Transformer-based models. L: Number of hidden layers; H: Hidden size; S: Sequence length.}
% \label{tab:appendix-details-for-five-transformer-based-models}
% \vskip 0.15in
% \begin{center}
% \begin{small}
% \begin{sc}
%     \begin{tabular}{ccccc}
%     \toprule
%     Model & L & H & S & \#params \\
%     \midrule
%     BERT-Huge & 32    & 1280  & 512   & 672M \\
%     T5-Large & 24/24 & 1024  & 512   & 737M \\
%     ViT-Huge & 32    & 1280  & 196 & 632M \\
%     Swin-Huge & 2/2/42/2 & 320/640/1280/2560 & 49*64/49*16/49*4/49*1 & 1.02B \\
%     Llama-7B & 32 & 4096 & 2048 & 7B \\
%     \bottomrule
%     \end{tabular}
% \end{sc}
% \end{small}
% \end{center}
% \vskip -0.1in
% \end{table}
\begin{table}[t]
    \begin{center}
    \begin{small}
	\begin{threeparttable}
\begin{tabular}{ccccccc}
    \toprule
    \multirow{2}{*}{Model} & \multirow{2}{*}{Task\tnote{1)}} & \multicolumn{4}{c}{Statistics} & \multirow{2}{*}{Precision} \\
    \cmidrule(r){3-6}
    && \#hidden layers  &  Hidden size   & Sequence length   & \#params &\\
    \midrule
    BERT-Huge & PT   & 32 & 1280 & 512 & 672M     & FP32      \\
    T5-Large  & CG   & 24/24 & 1024 & 512 & 737M    & FP32      \\
    ViT-Huge  & IC   & 32 & 1280 & 196 & 632M    & FP32      \\
    Swin-Huge & IC   & 2/2/42/2 & 320 & 49 $\times$ 64 & 1.02B   & FP32      \\
    Llama-7B  & CLM  & 32& 4096 & 2048 & 7B      & FP16   \\
    Llama-13B & CLM & 40 & 5120 & 2048 & 13B & FP16 \\
    \bottomrule
\end{tabular}
\begin{tablenotes}
    \footnotesize
    \item[1)] PT: Pretraining; CG: Conditional Generation; IC: Image Classification; CLM: Causal Language Modeling.
\end{tablenotes}
\end{threeparttable}
\end{small}
\end{center}
\vskip -0.1in
     \caption{Summary of the evaluated models.}
     \label{tab:summary-models}
\end{table}

\noindent\textbf{Model detail}\quad Table~\ref{tab:summary-models} summarizes six Transformer-based models selected for our evaluations. Four of these models, namely BERT-Huge~\citep{devlin_bert_2019}, T5-Large~\citep{raffel_exploring_2020}, Llama-7B, and Llama-13B~\citep{touvron_llama_2023,touvron_llama_2023-1}, belong to the domain of natural language processing (NLP). At the same time, the remaining two, ViT-Huge~\citep{dosovitskiy_image_2021} and Swin-Huge~\citep{liu_swin_2021}, are associated with computer vision (CV). It is noteworthy that BERT, ViT, and Llama maintain consistent types of hidden layers respectively, whereas T5 and Swin have different types of hidden layers. Numbers separated by slashes represent the statistical information for different layer types. For instance, Swin-Huge comprises four types of layers, each with 2, 2, 42, and 2 layers, respectively.

\noindent\textbf{Training detail}\quad UniAP is based on the PyTorch framework and integrates models from HuggingFace Transformers. It employs various types of parallelism, including Pipeline Parallelism~(PP), Data Parallelism~(DP), Tensor Parallelism~(TP), and Fully Sharded Data Parallelism~(FSDP), utilizing GPipe~\citep{huang_gpipe_2019}, PyTorch DDP~\citep{li_pytorch_2020}, Megatron-LM~\citep{narayanan_efficient_2021}, and FairScale~\citep{fairscale_authors_fairscale_2021}, respectively. For NLP models, we use the English Wikipedia dataset~\citep{wikidump}, while the ImageNet-1K dataset~\citep{imagenet15russakovsky} is used for CV models. We train these models using the Adam optimizer~\citep{kingma_adam_2015}. We omit hyperparameters here such as learning rate and weight decay as these have minimal impact on training throughput. 
The model parameters in the HuggingFace Transformers are configured to align with the specifications of each individual model. For instance, we set \textit{hidden\_size} to 1280, \textit{num\_hidden\_layers} to 32, \textit{num\_attention\_heads} to 16, and \textit{seq\_length} to 512 for BERT-Huge. Regarding other hyperparameters in the HuggingFace configurations, we set \textit{hidden\_dropout\_prob} and \textit{attention\_probs\_dropout\_prob} to 0.0 for ViT-Huge. For Swin-Huge, we set \textit{drop\_path\_rate} to 0.2. We remain other configurations to default. It should be noted that the training batch sizes for each experiment are outlined in the main paper.

% \section{Scalability}\label{appendix:experiment:scalability}
% \begin{figure*}[t]
% \begin{center}
% \vskip 0.2in
% \begin{subfigure}{0.47\linewidth}
% \centering
% \includegraphics[width=\linewidth]{./scalability_proportional.pdf}
% \label{fig:scalability:throughput}
% \end{subfigure}
% \begin{subfigure}{0.47\linewidth}
% \centering
% \includegraphics[width=\linewidth]{./scalability_time_proportional.pdf}
% \label{fig:scalability:optimization-time}
% \end{subfigure}
% \vskip -0.2in
% \caption{Training throughput and strategy optimization time with different number of nodes for different models. $n_{nodes}$ denotes the number of nodes~(machines) and $bsz$ denotes the mini-batch size. \textit{Left:} Training throughput. \textit{Right:} Strategy optimization time.}
% \label{fig:scalability}
% \end{center}
% \end{figure*}
% We study the scalability of UniAP on \textsc{EnvD}, the result of which is shown in Figure~\ref{fig:scalability}. We can find that the training throughput of the optimal strategy and the corresponding strategy optimization time demonstrate near-linearity as the number of nodes and mini-batch size increase. This phenomenon reveals that UniAP is scalable and verifies the computational complexity analysis in Section~\ref{subsec:method:unified-optimization}.

\section{Estimation accuracy}\label{appendix:experiment:performance_modeling}
Some variables in UniAP and other AP methods are estimated values rather than actual running values. The TPI~(inverse of training throughput) returned by UniAP and other AP methods is one of them. Accurate estimation for TPI or training throughput is crucial for evaluating candidate parallel strategies and ensuring the optimality of the solution. To quantify the accuracy of the estimated training throughput, we introduce a metric called \emph{relative estimation error~(REE)} $e$ for training throughput:
\begin{equation}
    e(T, \hat{T}) = \frac{|T - \hat{T}|}{T} \times 100\%,
\end{equation}
where $T$ is the actual training throughput and $\hat{T}$ is the estimated training throughput.

\begin{figure}
\begin{center} 
\centering
\includegraphics[width=0.65\linewidth]{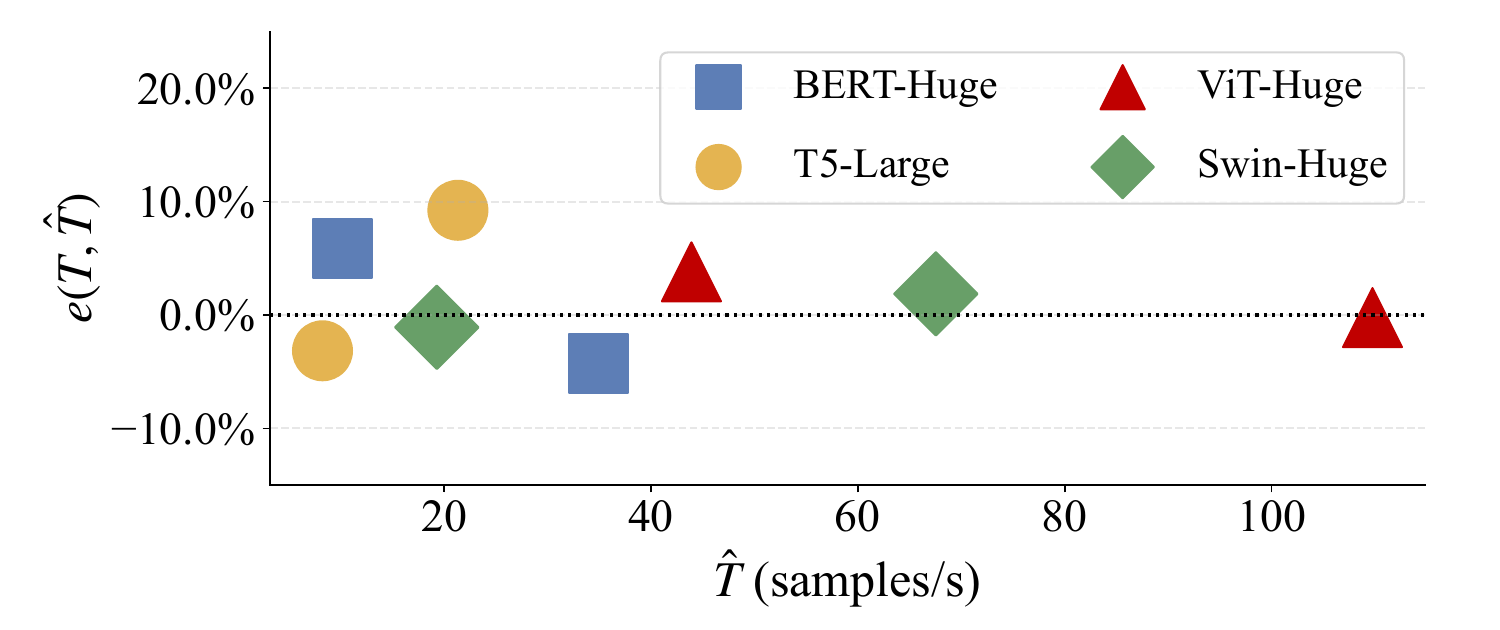}
\caption{Relative estimation error.}
\label{fig:performance-modeling}
\end{center}
\end{figure}

We evaluate the optimal parallel strategies obtained from \textsc{EnvA} and \textsc{EnvB} and visualize the REE of UniAP in Figure~\ref{fig:performance-modeling}. The results show that UniAP achieves an average REE of 3.59\%, which is relatively small. In contrast, the average REE for Galvatron~\citep{miao_galvatron_2022} in our experiments is 11.17\%, which is larger than that of UniAP.

\section{Case study: BERT-Huge}\label{appendix:case-study}
\begin{figure*}
\vskip 0.2in
\begin{center}
    \centering
    \includegraphics[width=0.9\linewidth]{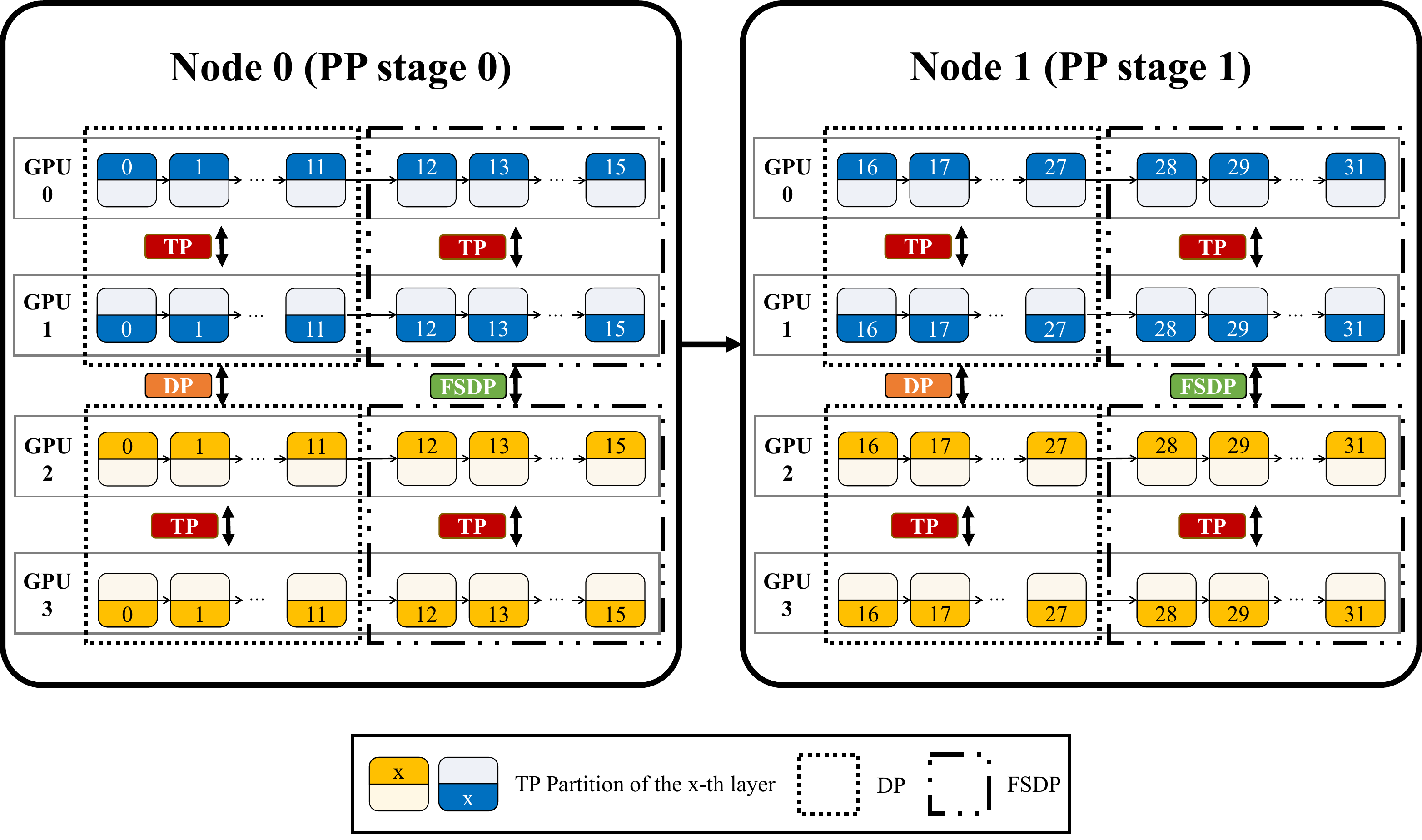}
    \caption{The optimal parallel strategy for all hidden layers of BERT-Huge on \textit{EnvB}. Different colors represent different input samples in a micro-batch.}
    \label{fig:visualization-model}
\end{center}
\vskip -0.2in
\end{figure*}
In this section, we present a visualization of the optimal parallel strategy discovered by UniAP. As represented in Figure \ref{fig:visualization-model}, the strategy pertains to training BERT-Huge with 32 hidden layers in a 2-node environment \textit{EnvB} with a mini-batch size of 16. Each node was equipped with 2 Xeon E5-2620 v4 CPUs, 4 TITAN Xp 12GB GPUs, and 125GB memory. These nodes are interconnected via a 10Gbps network. It should be noted that we only showcase the parallel strategy for the hidden layers here for simplicity but without losing generality.

\begin{figure}[t]
\vskip 0.2in
\begin{center}
    \centering
    \includegraphics[width=0.48\linewidth]{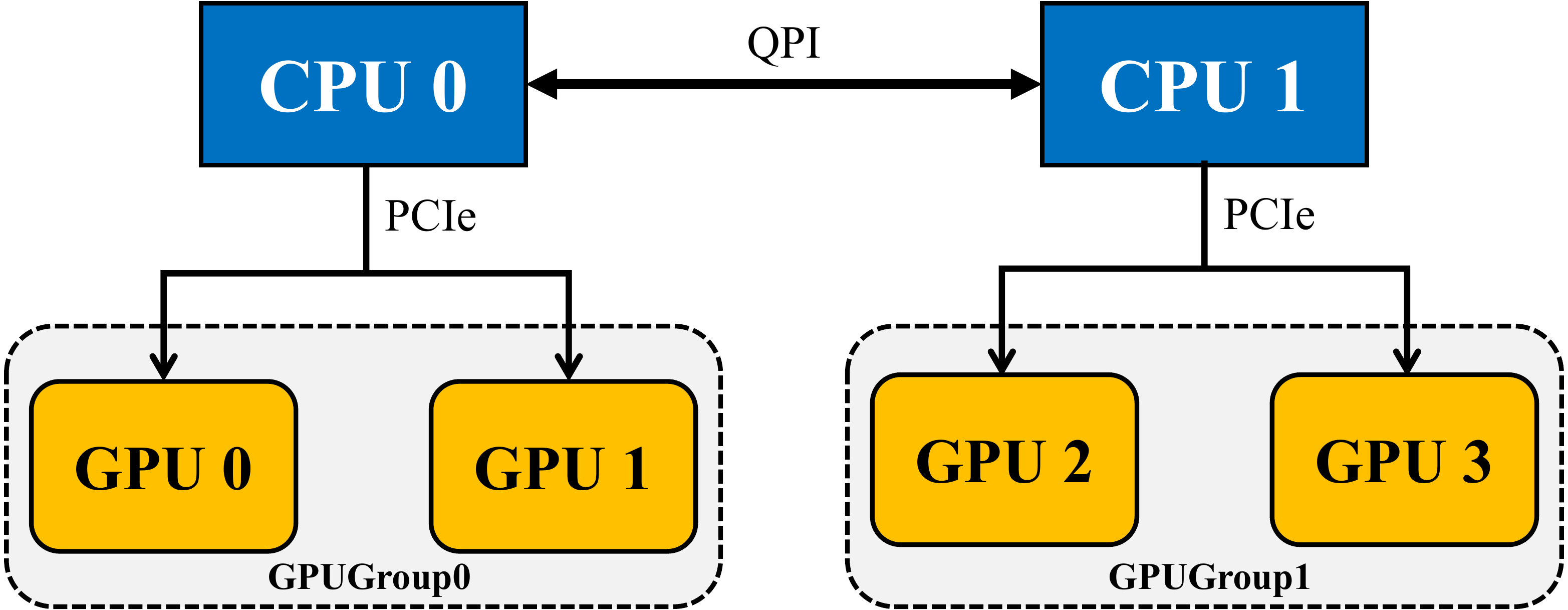}
    \caption{Topology of a node in \textit{EnvB}.}
    \label{fig:topo}
\end{center}
\vskip -0.2in
\end{figure}

Here, we provide further topological information for a node within \textit{EnvB}. As illustrated in Figure~\ref{fig:topo}, we categorize the GPUs numbered 0 and 1 in each node and refer to them collectively as \textit{GPUGroup0}. Similarly, we label the GPUs numbered 2 and 3 as \textit{GPUGroup1}. In \textit{EnvB}, the interconnects within each GPU group (i.e., PCIe) have superior bandwidth than that between different groups (i.e., QPI). We collectively designate these two connection bandwidths as intra-node bandwidth, which is higher than inter-node bandwidth.

In this example, UniAP has identified a parallel strategy for inter-layer parallelism that involves a two-stage pipeline. This strategy utilizes parallelism in a manner that is both efficient and effective. Specifically, the communication cost of point-to-point~(P2P) between two nodes is less than that of all-reduce. Given that the inter-node bandwidth is lower than the intra-node bandwidth, the two-stage PP becomes a reasonable choice. Moreover, the pipeline has been designed such that each stage comprises an equal number of layers. This design leverages the homogeneity of the nodes and ensures load balancing across the cluster.

Within each PP stage, UniAP employs an intra-layer parallel strategy. It utilizes a 2-way DP for the initial 12 hidden layers in each stage between \textit{GPUGroup0} and \textit{GPUGroup1}. For the remaining four hidden layers, a 2-way FSDP is utilized between \textit{GPUGroup0} and \textit{GPUGroup1} to reduce memory footprint and meet memory constraints. Within each GPU group, UniAP employs a 2-way TP for each layer. In general, TP incurs more significant communication volumes than DP and FSDP. In order to achieve maximum training throughput on \textit{EnvB}, it is necessary to implement parallel strategies that prioritize higher communication volumes within each group and lower volumes between groups. Therefore, the strategy for BERT-Huge with 32 hidden layers combines the best elements of PP, DP, TP, and FSDP to maximize training throughput.

In addition, we have conducted calculations for the model FLOPs utilizatio~(MFU)~\citep{chowdhery_palm_2023} for Galvatron, Alpa, and UniAP in this scenario to validate our analysis. MFU is independent of hardware, frameworks, or implementations. Therefore, it allows us to examine the performance of different parallel strategies solely from a strategic perspective. For BERT-Huge, the resulting MFUs for UniAP, Galvatron, and Alpa are 58.44\%, 58.44\%, and 55.10\% on \textsc{EnvA}, while 23.6\%, 13.7\%, and 19.6\% on \textsc{EnvB}, respectively. These results validate that UniAP's optimization of inter- and intra-layer AP results in superior performance compared to Galvatron and Alpa.

\section{Limitation}\label{appendix:limitation}
UniAP is currently designed and tested on homogeneous clusters, but incorporating automatic parallelism for training deep models on heterogeneous clusters~(e.g., a cluster equipped with both NVIDIA GPUs and DCUs) is another important research topic. Given that current parallel techniques primarily target homogeneous clusters with limited emphasis on heterogeneous clusters, we have chosen to leave this topic for future exploration.